\documentclass[conference]{IEEEtran}
\IEEEoverridecommandlockouts
\usepackage{cite}
\usepackage{caption}
\usepackage{amsmath,amssymb,amsfonts}
\usepackage{algorithmic}
\usepackage{graphicx}
\usepackage{textcomp}
\usepackage{xcolor}
\usepackage{dblfloatfix}
\usepackage{url}

\def\BibTeX{{\rm B\kern-.05em{\sc i\kern-.025em b}\kern-.08em
    T\kern-.1667em\lower.7ex\hbox{E}\kern-.125emX}}

\title{LoopTune: Optimizing Tensor Computations with Reinforcement Learning\\
}
\begin{document}

% \author{\IEEEauthorblockN{Anonymous Authors}}

\author{\IEEEauthorblockN{Dejan Grubisic}
\IEEEauthorblockA{\textit{Rice University, Meta AI} \\
\textit{Houston, TX, USA}}
\and
\IEEEauthorblockN{Bram Wasti}
\IEEEauthorblockA{\textit{Meta AI} \\
\textit{New York, NY, USA}}
\and
\IEEEauthorblockN{Chris Cummins}
\IEEEauthorblockA{\textit{Meta AI} \\
\textit{Menlo Park, CA, USA}}
\and
\IEEEauthorblockN{John Mellor-Crummey}
\IEEEauthorblockA{\textit{Rice University} \\
\textit{Houston, TX, USA}}
\and
\IEEEauthorblockN{Aleksandar Zlateski}
\IEEEauthorblockA{\textit{Meta AI, Dabun AI} \\
\textit{New York, NY, USA}}
}

\maketitle

% TODO:
% How is LoopTune different from LoopNest/LoopTool?
% Can you use larger matrices 512x3072x768?
% 64 schedules is not enough for AutoTVM -> 2k is more appropriate.
% Why APEX\_DQN performs much better than other RL algorithms?
% What is the hyperparameters to train the APEX-DQN? Are the hyperparameters the same in all benchmarks?

% Related work is not about tensor computation but frameworks that optimize them!

\vspace{-0.4in}
\begin{abstract}
Advanced compiler technology is crucial for enabling machine learning applications to run on novel hardware, but traditional compilers fail to deliver performance, popular auto-tuners have long search times and expert-optimized libraries introduce unsustainable costs. To address this, we developed LoopTune, a deep reinforcement learning compiler that optimizes tensor computations in deep learning models for the CPU. LoopTune optimizes tensor traversal order while using the ultra-fast lightweight code generator LoopNest to perform hardware-specific optimizations. With a novel graph-based representation and action space, LoopTune speeds up LoopNest by 3.2x, generating an order of magnitude faster code than TVM, 2.8x faster than MetaSchedule, and 1.08x faster than AutoTVM, consistently performing at the level of the hand-tuned library Numpy. Moreover, LoopTune tunes code in order of seconds.

\end{abstract}
\vspace{1\baselineskip}

\newcommand\blfootnote[1]{%
  \begingroup
  \renewcommand\thefootnote{}\footnotetext{#1}%
  \addtocounter{footnote}{-1}%
  \endgroup
}

% \vspace{1\baselineskip}

\section{Introduction}
\blfootnote{Correspondence to Dejan Grubisic on dejan.grubisic@rice.edu}
Contemporary advances in machine learning (ML) have led chip designers to develop a plethora of extremely powerful chips to accommodate computationally intensive ML workloads. For instance, Nvidia introduced tensor cores \cite{markidis2018nvidia, choquette2021nvidia}, Intel and AMD added specialized AVX \cite{lomont2011introduction, jeong2012performance}, FMA \cite{wittmann2015short}, and VNNI instructions \cite{tekin2021state}, while  Google introduced TPUs \cite{jouppi2017datacenter}. Moreover, hardware companies started making ML-specific chips such as Graphcore \cite{jia2019graphcore} and Cerebras \cite{rocki2020cerebras}. 

To fully harness the power of advanced hardware, advanced compiler technology is a must. However, traditional compilers have several limitations that impede their ability to do so.

First, traditional compilers have been developed for a limited set of ISAs with similar programming models, making it difficult to adapt them to exotic hardware with different chip resources. Even with the front/back-end separation introduced by LLVM, the task remains challenging, because traditional representations are not easily optimized for novel hardware.

Second, as traditional compilers are extended to cover more use cases, they become increasingly complex, with hundreds of optimization passes that frequently depend on one another. This complexity increases development and maintenance cost.

Finally, traditional ``catch-all" compiler techniques fail to fully utilize novel resources available on emerging hardware designed for specific workloads.

\begin{figure}[ht]
% \vskip 0.2in
\begin{center}
\centerline{\includegraphics[width=\columnwidth]{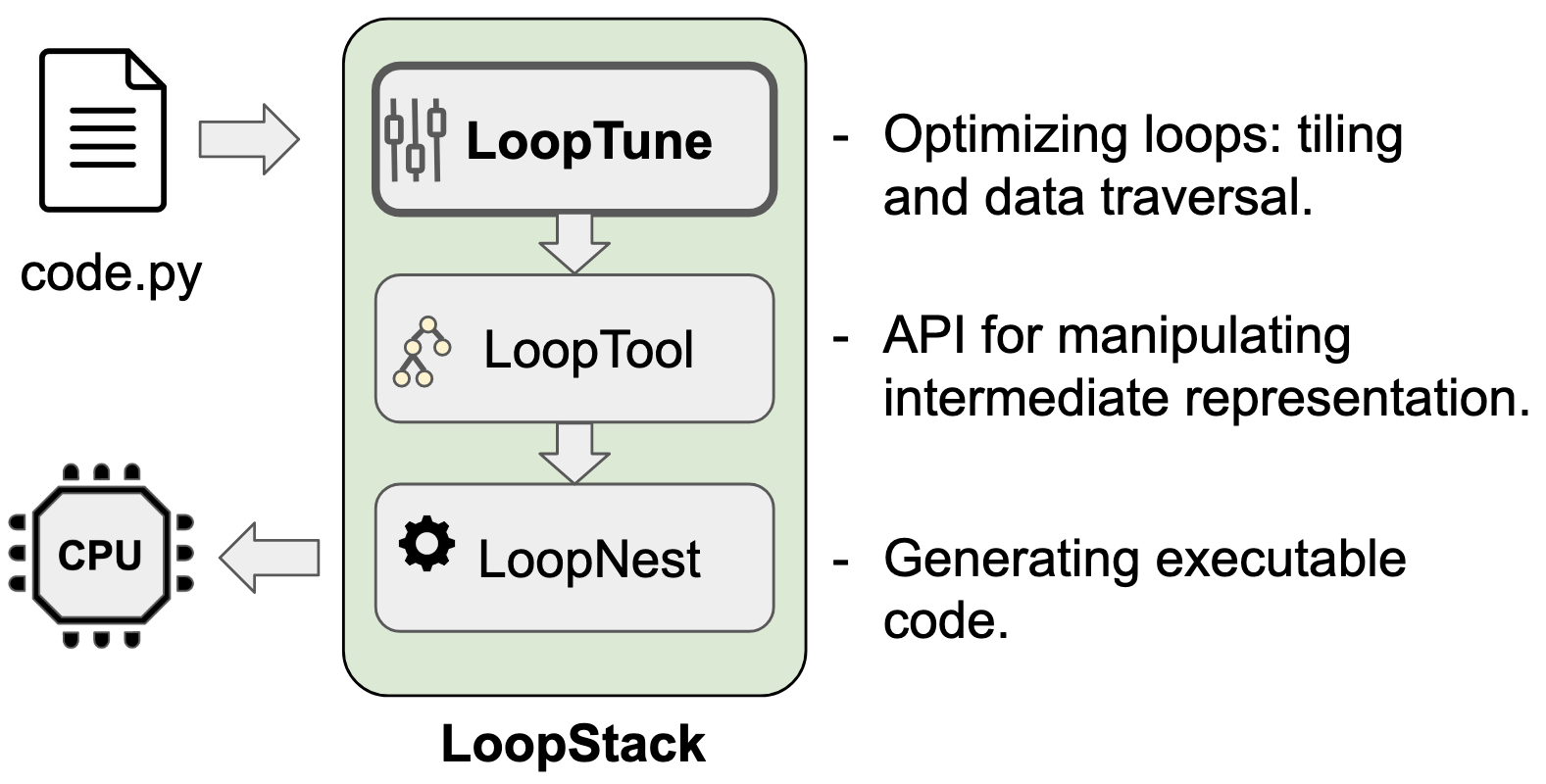}}
\caption{LoopStack architecture.}
\label{fig/loopstack}
\end{center}
\vskip -0.4in   
\end{figure}

\begin{figure*}[ht]
\begin{center}
\centerline{\includegraphics[width=0.8\textwidth]{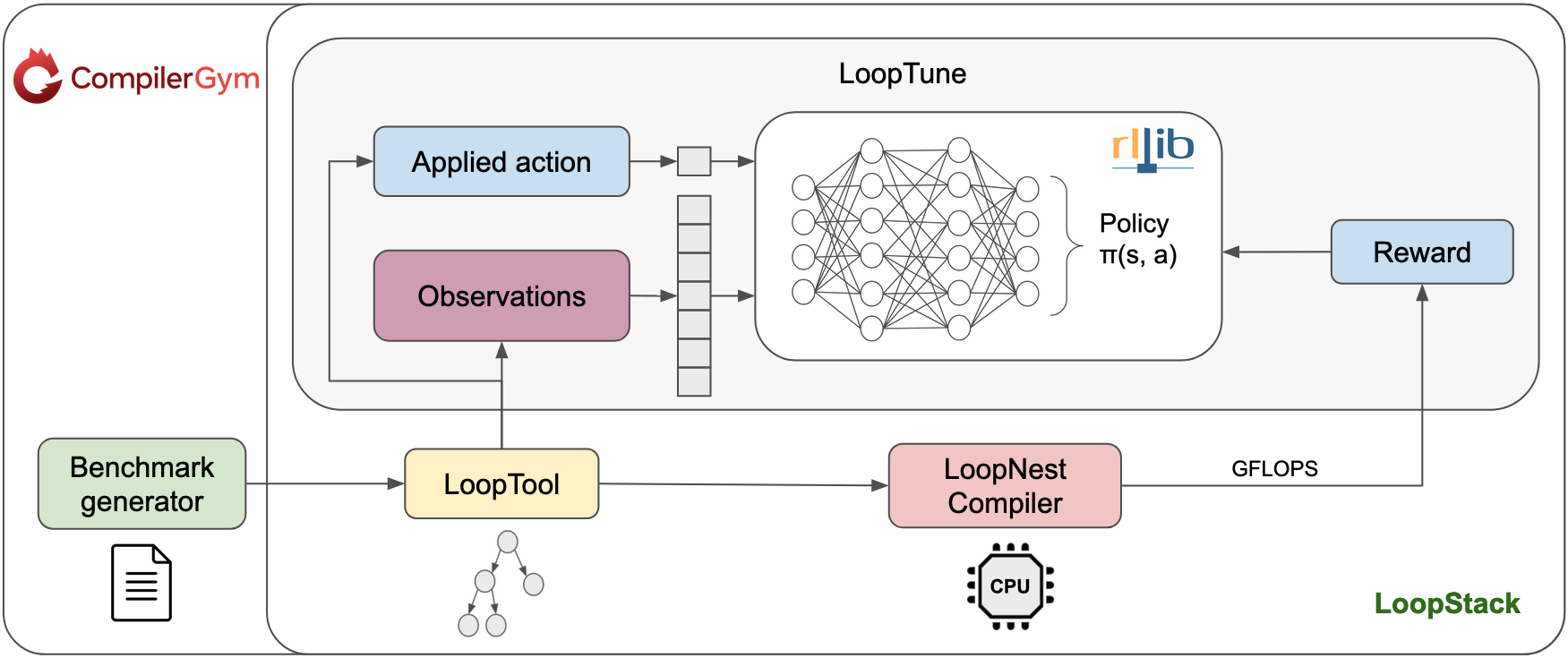}}
\caption{LoopTune training loop. LoopTune transforms generated benchmark to an intermediate representation (IR), and uses LoopTool API to apply actions and get observations, while LoopNest compiles and executes the loop nest providing the reward.}
\label{fig/framework}
\end{center}
\vskip -0.3in
\end{figure*}

So, what are our alternatives to traditional compilers? Expert-optimized libraries, auto-tuners, or something else? 

Expert-optimized libraries require an enormous time investment by experts and the work must be repeated for each new device.
 High-performance tensor operation libraries such as cuDNN \cite{chetlur2014cudnn}, OneDNN \cite{onednn}, or XNNPACK \cite{xnnpack} are usually tied to a narrow range of hardware devices and tend to be large in code size, which may impede their use on mobile devices.

As an alternative approach, projects like Halide \cite{ragan2013halide} and TVM \cite{tvm} optimize a high-level representation of a loop nest performing a tensor computation with a discrete set of transformations such as loop reordering and tiling before compiling it to a particular target hardware with the LLVM compiler. This approach provides high performance and eliminates the need for expert-optimized libraries, but introduces an astronomical number of possible schedules. 

To optimize a simple problem, such as Local Laplacian Filters, Halide estimates a lower bound of \(10^{720}\) possible schedules \cite{ragan2013halide}. To find performant schedules in such a huge space,
Halide and TVM use genetic algorithms and parallel simulated annealing with a trained cost model \cite{tvm} respectively. Both approaches suffer from very large compilation times.

Contemporary breakthroughs in deep reinforcement learning (deep RL) in complex video games, such as those in Atari \cite{mnih2013playing}, and AlphaGo \cite{silver2016mastering}, have inspired the compiler research communities to attempt to leverage deep RL as well \cite{haj2019autophase, haj2020neurovectorizer, brauckmann2021reinforcement, wang2022automating}. Similar to iterative algorithms, the deep RL agent explores an optimization space. However, there is one important difference - knowledge of the search space is embedded into a neural network. Inferring the neural network then replaces part of the search for optimizations. This enables example-driven, fast optimization-space search is exactly what ML-specific, as well as general compilers, need.  

In our work, we further build on recent RL-based efforts in compiler research by extending LoopStack \cite{wasti2022loopstack} - an ultra-fast tensor compiler backend, with LoopTune to find a performant loop schedule with deep reinforcement learning (Figure \ref{fig/loopstack}). LoopTune manipulates loop schedule space through an API called LoopTool and generates the binary from the given schedule with LoopNest. The goal of LoopTune is to train the policy network that will find a close-to-optimal schedule for the given loop nest in a few steps, decreasing auto-tuning time to the order of seconds.

In this paper, we present the following novel contributions:
\begin{itemize}
    \vspace*{0.05in}
    \item LoopTune - a deep RL framework that finds performant loop ranges and schedules.
    \vspace*{0.05in}
    \item A novel graph-based embedding of tensor computations suitable for reinforcement learning.
    \vspace*{0.05in}
    \item A novel action space for tunning tensor computation suitable for RL training.
    \item Comparison of 5 popular RL algorithms in optimizing loop schedule.
    \vspace*{0.1in}
\end{itemize}

By combining reinforcement learning with appropriate representations and a well-chosen optimizer, we are able to generate faster code than baseline traditional search techniques, outperform popular auto-tuners such as autoTVM and MetaSchedule, and perform at the level of an expert-optimized library Numpy. 
% \vspace{-0.2\baselineskip}
\section{Background}
\begin{figure*}[ht!]
\setcounter{figure}{2}
\includegraphics[width=\textwidth]{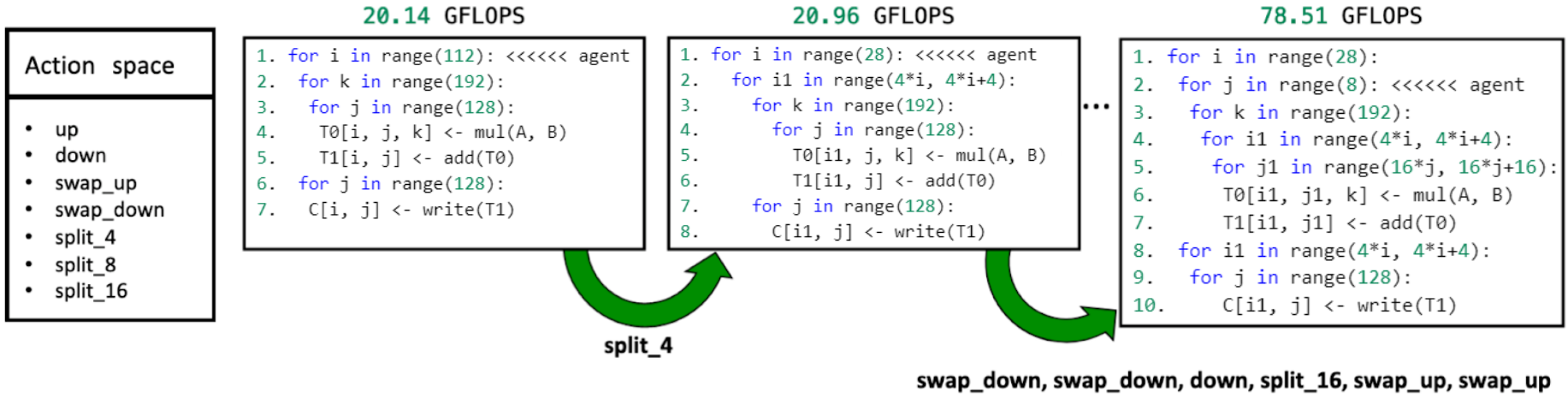}
\vspace*{-0.2in}
\caption{Optimizing ranges and order of loops for matrix multiplication using LoopTune's action space.}
\label{fig/optimizing_lt}
\end{figure*}

The principal component of machine learning workloads can be expressed as a series of tensor contractions. Tensor contractions represent the generalization of matrix multiplication, trace, transpose, and other commonly used operations on matrices to higher dimensions. 

Formally, we can define tensor contractions in the following way \cite{matthews2018high}. Let \( \mathcal{A, B, C} \) be tensors with dimensions of \( d_A, d_B, d_C \) respectively. Similar to 2D matrix multiplication, for each pair of tensors \( \mathcal{AB, AC, BC} \) we define the dimensions both tensors will iterate together. Namely, these indices will have dimensions \( I_{AB} = (d_A + d_B - d_C) / 2 \), \( I_{AC} = (d_A + d_C - d_B) / 2 \) and \( I_{BC} = (d_B + d_C - d_A) / 2 \). Then, tensor contraction can be defined with:

\begin{center}
  \begin{minipage}{\linewidth}
    \centering
    \includegraphics[width=0.9\textwidth]{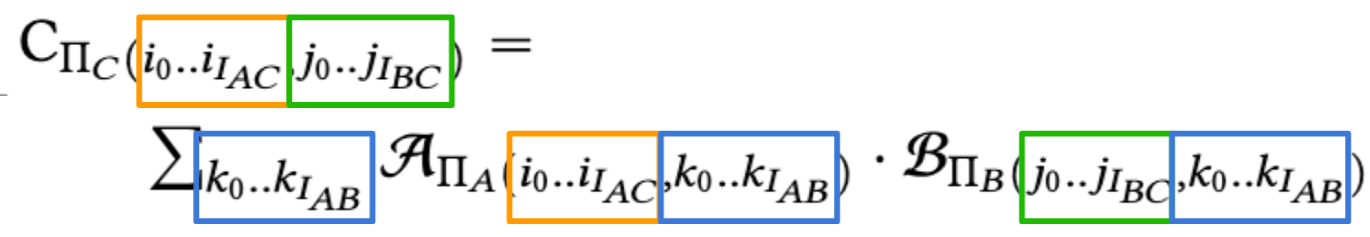}
    \setcounter{figure}{0}
    % \captionof{figure}{LoopStack architecture.}
  \end{minipage}
\end{center}
where \(\cdot\) is scalar multiplication and \(\Pi\) stands for all permutations of specified dimensions. Note that here we have to do all permutations to keep the result consistent, since the iterating dimensions may be chosen in any order. To simplify notation further, we can use Einstein notation and implicitly sum over dimensions that don't exist in the resulting tensor.
\[ 
    \mathcal{C}_{\Pi_C(I, J)} =
    \mathcal{A}_{\Pi_A(I, K)}
    \cdot
    \mathcal{B}_{\Pi_B(J, K)}
\]
To allow the use of the non-linear activation function used in deep learning, we extend our notation with an element-wise operation that transforms the final result (post).
\[ 
    \mathcal{C}_{\Pi_C(I, J)} =
    post(
    \mathcal{A}_{\Pi_A(I, K)}
    \cdot
    \mathcal{B}_{\Pi_B(J, K)}
    )
\]

With these extensions, we can express not only general matrix-to-matrix multiplication (GEMM), matrix-to-vector multiplication (GEMV), and vector-to-matrix multiplication (GEVM) operations, but also general machine learning primitives such as:

\begin{itemize}
    % \vspace*{-0.1in}
    \item Convolutions \cite{krizhevsky2017imagenet} : \:\:\:\:\:\:
    \( O_{R,C} = I_{R+K, C+J} \cdot \omega_{K,J} \)
    \vspace*{0.05in}
    \item Pooling \cite{krizhevsky2017imagenet} : \:\:\:\:\:\:\:\:\:\:\:\:\:\:\:\:
    \( O_{R,C} = max(I_{2R, 2C} ) \)
    \vspace*{0.05in}
    \item Reductions \cite{abdi2010pca} : \:\:\:\:\:\:\:\:\:\:\:\:
    \( O_{R} = I_{R,C} \)
    \vspace*{0.05in}
    \item Transpositions \cite{abdi2010pca} : \:\:\:\:\:\:
    \( O_{R,C} = I_{C, R} \)
    \vspace*{0.05in}
    \item Concatenations \cite{radu2018multimodal} : \:\:\:
    % \( O_{R,C_1} = A_{R, C_1} ; O_{R,C_2} = B_{R, C_2}  \)
    \( O_{R,C_1+C_2} = A_{R,C_1}|B_{R,C_2}\)
    \vspace*{0.05in}
    \item Broadcast \cite{albooyeh2019broadcast} : \:\:\:\:\:\:\:\:\:\:\:\:\:\:\:
    \( O_{R,C} = I_{R} \)
    \vspace*{0.1in}
\end{itemize}

Besides machine learning, tensor contractions are widely used in physics simulations, spectral element methods, quantum chemistry, and other fields. Despite many efforts \cite{grosser2012polly, di2014towards, matthews2018high}, none of the state-of-the-art production compilers such as GCC \cite{stallman1999using} and LLVM \cite{lattner2004llvm} can automatically transform naive tensor contraction loop nests to expertly-tuned implementations.

\section{Learning to Optimize Tensor Programs}
To optimize tensor operations, we separate the problem of finding optimal loop range and order (schedule) from hardware-dependent low-level optimizations, such as vectorization.
To find performant schedules LoopTune uses deep reinforcement learning to train a policy network, while LoopNest \cite{wasti2022loopstack} applies low-level tensor optimizations and generates executable code given a schedule. 
%We provide more details about LoopNest in Appendix \ref{sec/loop_nest}. 
LoopTune could be also connected to all backends that optimize a given loop schedule such as Halide and TVM, which could be an interesting future direction.

Figure \ref{fig/framework} outlines the workflow of LoopTune. The process begins by creating an RL environment by using CompilerGym \cite{cummins2022compilergym}. This framework allows us to map the problem of finding performant schedules to RL methodology and using state-of-the-art libraries such as RLlib \cite{liang2018rllib} for training. To create an environment in CompilerGym we define an action space, an observation space, and a reward that will be used as an optimization criterion during RL training. Additionally, we define a set of benchmarks that represent tensor operations. For all benchmarks, we assume that loop bounds are constant.

In each training epoch, we convert the benchmark to an intermediate representation by adding an ``agent" annotation to the first loop (Figure \ref{fig/optimizing_lt}). In each step, the agent applies an action from the action space to the current loop changing the loop schedule. LoopTune encodes our novel graph-based state representation to a vector representation (described in Section \ref{sec:representation}) and feeds it to the reinforcement learning training loop. Finally, LoopNest compiles and evaluates the loop schedule which provides a reward signal for training the policy network. 

In the inference phase, LoopTune iteratively calculates the best action by the policy network and applies it to the current state. Since this procedure doesn't include loop nest evaluation it is fast and constrained only to the speed of the inference. Practically, this enables the policy network to quickly reach the desired state in a matter of seconds.
\vspace{-0.01in}

\begin{figure*}
\setcounter{figure}{3}
\includegraphics[width=\textwidth]{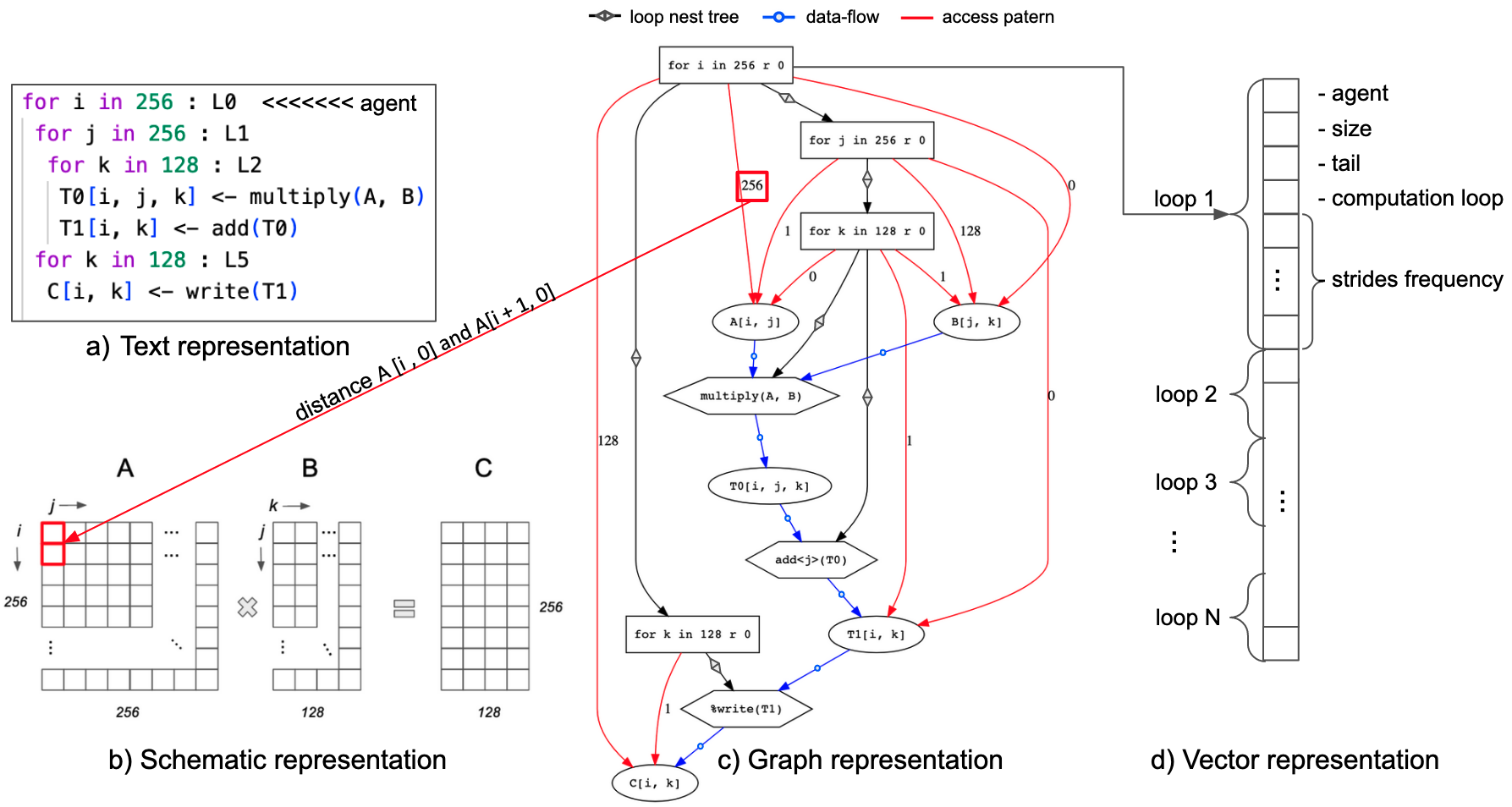}
\vspace*{-0.2in}
% \caption{Novel graph encoding for tensor operations.}
\caption{Text representation shows the algorithm. Schematic representation shows the memory layout. Graph representation explains nesting order (black), access pattern (\textcolor{red}{red}), and data flow (\textcolor{blue}{blue}). Vector representation aggregates graph representation for the training.}
\label{fig/mm_graph}
\vspace*{-0.2in}
\end{figure*}

\subsection{Defining an Action Space}\label{sec/action_space}

The LoopTool API \cite{wasti2022loopstack}, provides LoopTune with the ability to swap the positions of two loops, given their line numbers, and split a loop, given its line number and a specified tile size. Rather than having such parametric actions, that are inherently hard to train \cite{kanervisto2020action}, LoopTune defines a novel action space shown in Figure \ref{fig/optimizing_lt} and introduces the abstraction of an agent that traverses loop nests and applies actions on each loop. 

The agent uses \textit{up} and \textit{down} actions to move the cursor without changing the loop nest structure.
The \textit{swap\_up} and \textit{swap\_down} actions allow the agent to exchange the position of the current loop with its neighbor, moving the agent's cursor respectively. The \textit{split} family of actions creates a new loop with the same iterator, dividing the loop range with the specified split parameter. If the split parameter does not evenly divide the loop range, the current loop will have a remainder or ``tail", which will be executed at the end of the loop nest execution.

By limiting the action space in this way, we are able to simplify the problem in several ways. For example, a smaller number of possible actions enables the RL algorithm to explore and become more confident \cite{kanervisto2020action} with each action for different states. This might force the agent to use longer sequences of actions to reach certain states, but this is not a problem since each action other than \textit{up} and \textit{down} changes the loop nest and provides a non-zero reward signal. Furthermore, many states benefit from similar action sequences, which allows training to converge faster.

To keep the design simple, we decided to apply a fixed number of actions for optimization, rather than having an action that terminates the search. Our experiments have shown that having such an action often prevents exploration and converges to local minima. Instead, we rely on an implicit stop, which occurs when the agent starts oscillating between states that differ only by the cursor position.

\subsection{Defining a Reward}\label{sec/reward_function}

For the evaluation metric, we use billions of floating-point operations per second (GFLOPS). To measure GFLOPS, LoopTune uses LoopNest to compile and execute loops on a CPU. To ensure reliable results, LoopNest excludes the first 20 iterations as a warm-up and times multiple executions of the loop nest, taking the fastest measurement. 

During training, the agent applies action (A) from state S, moving it to the next state S'. The feature extractor maps the internal representation to the vector (S) which is used as input to the neural network. LoopNest calculates the reward for the applied action using the formula: 
\vspace*{0.01in}
\[ 
    reward = \frac{GFLOPS(S') - GFLOPS(S)}{GFLOPS\_PEAK\_PERFORMANCE}
\]
\vspace*{0.01in}

This normalizes all rewards, making training more stable. Rather than relying on peak performance from hardware specifications that may be imprecise, we evaluate peak performance empirically before the training by running the series of kernels with high arithmetic intensity, which always falls within a few percent of the theoretical peak. Finally, we send a tuple (S, S', A, R) to the RL library that performs one training step.

\subsection{Defining the State Representation}\label{sec:representation}
Each loop nest consists of a nest that computes operations and a write-back nest that writes the result to the memory. For state representation, we use the graph-based representation shown in Figure \ref{fig/mm_graph}. On this graph, there are 3 kinds of nodes: loops (rectangles), data (ellipses), and computation (diamonds). There are 3 kinds of edges. Black edges connect loops and computations that are nested from top to bottom. Blue edges represent data flow, while red edges represent the strides of each loop accessing tensors that read from memory (A, B), or write to memory (T). 
Stride is the distance in memory between two elements of a tensor when we increment only the index of a given loop. If this number is large, the iterating loop will try to fetch distant data in memory that may not be stored in the cache, resulting in a cache miss.

To make our representation usable for standard RL optimizers, we map the key features to a vector. In our vector representation, each loop is described with 20 integer values, namely:
\begin{itemize}
    \item (1) Is the agent's cursor on the loop
    \vspace*{0.05in}    
    \item (1) Loop size
    \vspace*{0.05in}
    \item (1) Loop tail
    \vspace*{0.05in}
    \item (1) Does loop belong to computation or write-back nest
    \vspace*{0.05in}
    \item (16) Histogram of strides frequency
\end{itemize}

\begin{figure}[ht]
% \vskip 0.2in
\begin{center}
\centerline{\includegraphics[width=0.75\columnwidth]{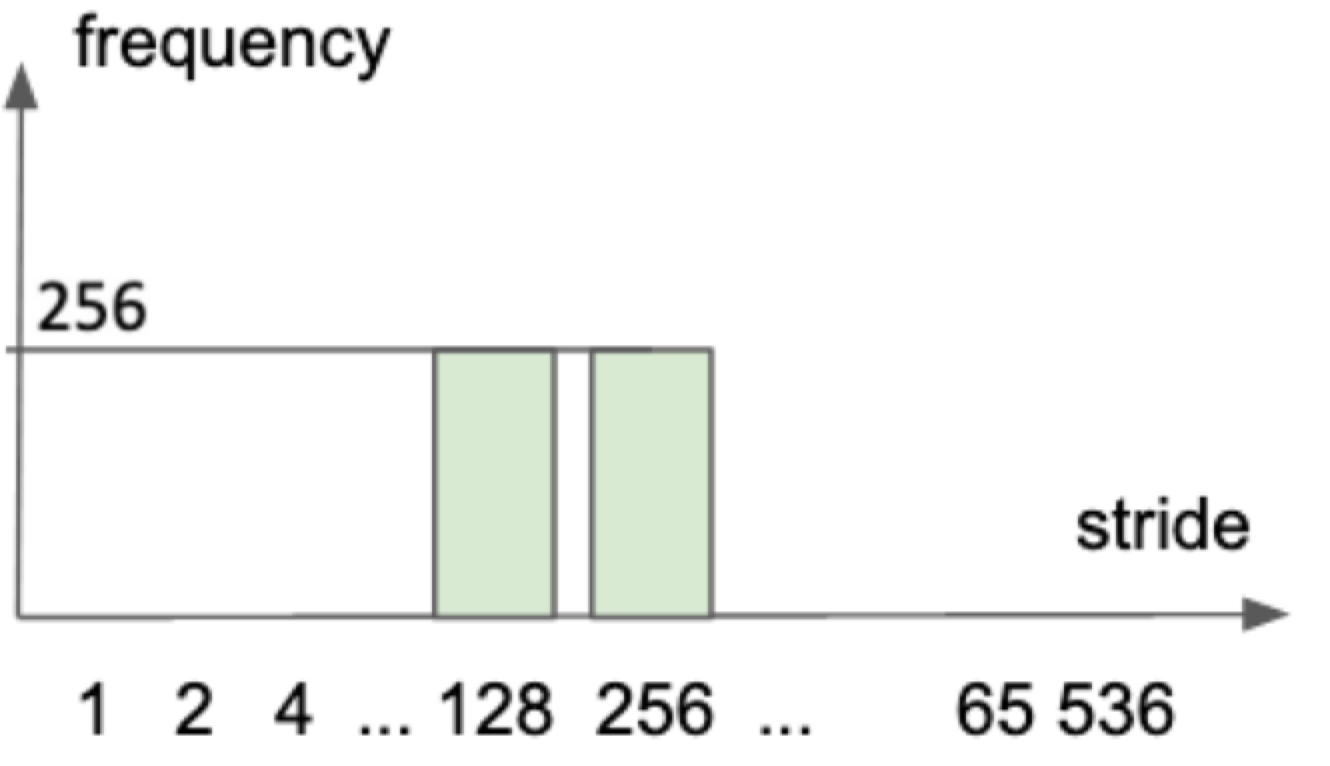}}
\caption{Histogram of strides frequency.}
\label{fig/stride_histogram}
\end{center}
\vskip -0.3in
\end{figure}

The histogram of strides frequency (Figure \ref{fig/stride_histogram}) represents the cumulative distribution of access strides for each loop. In other words, it shows how many accesses with given strides are produced from the given loop. For each loop, we calculate strides from the tensor shape and iterator position. Since stride can be an arbitrary integer larger than zero, we discretize strides to bins of size \(2^N\), where $N \in \{0...15\}$ to match the sizes of cache lines.

Agent bits are necessary since they give meaning to all actions since they depend on the cursor position. Size and tail bits define how many times memory is accessed with each stride, which distribution is captured in strides frequency. The computation loop shows whether the loop is used for computation or writeback.

We believe this is a minimal set of features for the RL algorithm to learn memory access patterns, which is key to optimizing for memory-bound computations such as tensor contractions. For applications that are compute-bound, adding features that describe computation in the loop body would be beneficial.

% Rllib library
\subsection{Library for Reinforcement Learning}
To optimize the training process of our reinforcement learning model, we have chosen to use RLlib \cite{liang2018rllib}, the performant library for reinforcement learning. In our work, we evaluate several learning algorithms, supported by RLlib, including Deep Q Learning (DQN), Apex Deep Q Learning (APEX\_DQN), Proximate Policy Optimization (PPO), Actor-Critic (A3C), and Impala.

DQN \cite{mnih2013dqn} attempts to learn the state value function by using experience replay for storing the episode steps in memory for off-policy learning, where samples are drawn from the replay memory at random. 

APEX\_DQN \cite{horgan2018dqnapex} creates instances of environment for each actor and collects the resulting experience in a shared experience replay memory prioritizing the most significant data generated by actors.

PPO \cite{schulman2017ppo} alternates between sampling data through the interaction with the environment while
using stochastic gradient ascent with minibatch updates.

A3C \cite{mnih2016a3c} calculates gradients on the workers directly in each episode and only broadcasts these gradients to the central model. Once the central model is updated parameters are sent back to the workers.

IMPALA \cite{espeholt2018impala} provides a scalable solution for collecting samples from individual agents and running stochastic gradient descent in the central loop.

To get out of the training choosing the suitable RL algorithm is crucial. Our experiments suggest that APEX\_DQN achieves converges significantly faster than other approaches that we elaborate on in the evaluation section. 
\section{LoopNest Backend Optimizer}\label{sec/loop_nest}

LoopNest \cite{wasti2022loopstack} is a powerful, domain-specific compiler that is specifically designed to optimize tensor programs with tuning with reinforcement learning in mind. Rather than relying on extensive general-purpose compilers, it utilizes a small set of expert-designed HPC optimizations, enabling fast code generation for each RL step. It includes custom primitives in code generation, custom assembly codes, instruction reordering, r-sum, and other optimizations suggested by optimization manuals for the target hardware \cite{arm2016arm, intel2014intel}.

Unlike traditional compilers, LoopNest takes into account user-defined orders of operations. This simplifies the code generator design and provides a more direct mapping between the quality of the user-defined order and its performance. This feature is particularly important for finding the best sequence of actions with reinforcement learning, and we firmly believe it makes the optimization space smoother and enables faster convergence.

Furthermore, LoopNest performs loop unrolling in a way that is consistent with hardware requirements and automatically vectorizes the innermost loop. It also applies register tiling \cite{domagala2014tiling}, keeping a portion of the output tensor in registers at all times. To reduce pressure on load/store units, LoopNest never generates code that spills registers to the stack, unlike traditional compilers like LLVM and GCC. It achieves this by finding the largest scope in which any modified values can fit in the register file.

When compared to LLVM \cite{lattner2004llvm}, which is commonly used as a backend choice for tensor compilers such as Halide \cite{ragan2013halide} and TVM \cite{tvm}, LoopNest achieves orders of magnitude faster compilation times while generating code that is equal or higher in performance on AMD (AMX512) architecture (Table \ref{fig/loopnest}). Halide is used to emit schedules for LLVM code generation, turning off runtime assertions and bound checks. 

\begin{table}[ht]
\caption{LoopNest vs. LLVM performance on AMD (AVX2) architecture \cite{wasti2022loopstack}.}
\vskip 0.2in
\begin{center}
\centerline{\includegraphics[width=1\columnwidth]{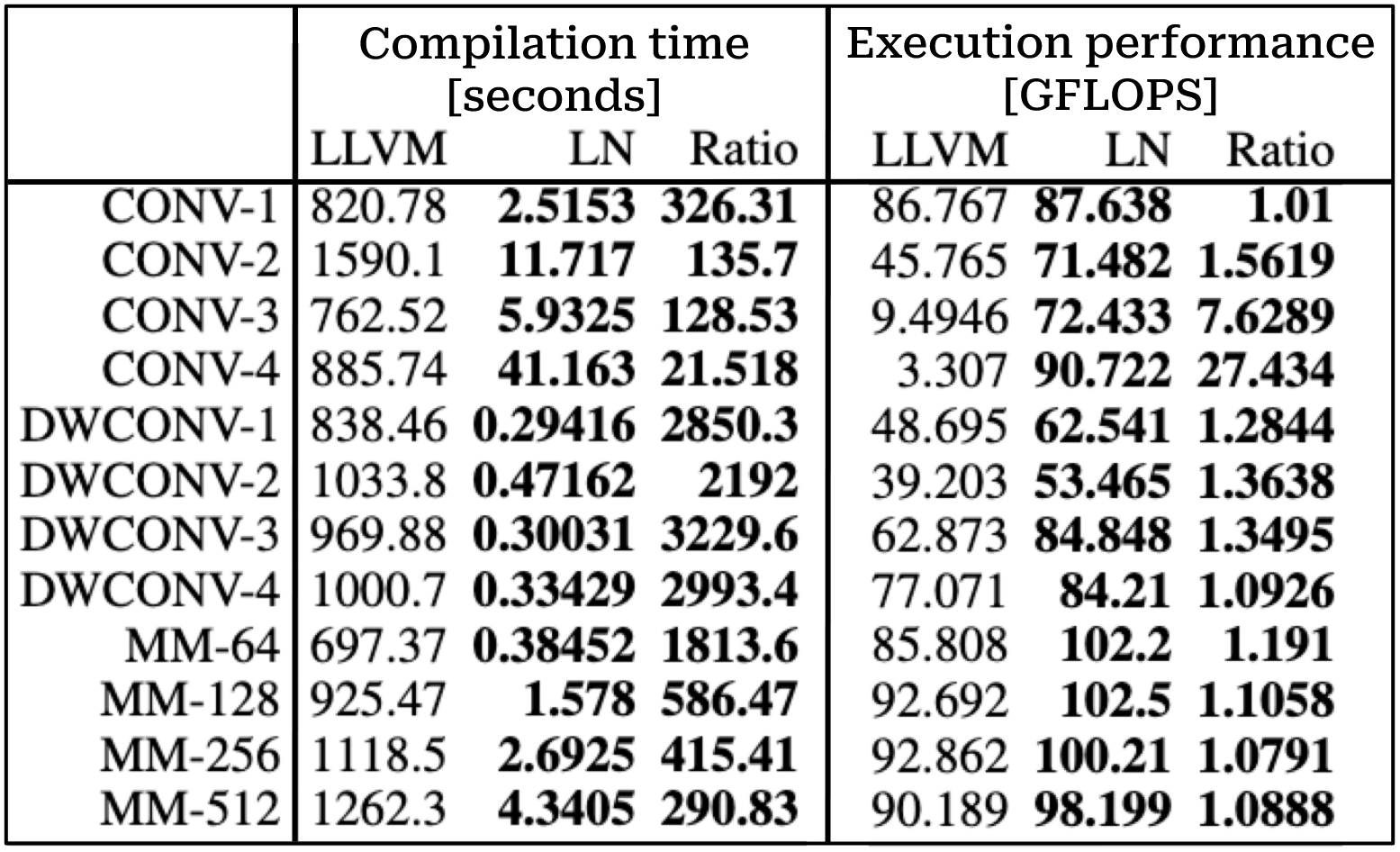}}
\label{fig/loopnest}
\end{center}
\vskip -0.2in
\end{table}

Beyond the AMD (AVX2) architecture, LoopNest achieves similar results for Intel (AVX512), Cortex A57, Cortex A73, NVIDIA Denver2, and Apple M1 architectures. Additionally, LoopNest has a small binary footprint of \(250Kb\), compared to LLVM's \(350Mb\) footprint, which makes LoopNest a compiler of choice for use on mobile and embedded systems.

\begin{figure*}[ht!]
\setcounter{figure}{5}
\includegraphics[width=\textwidth]{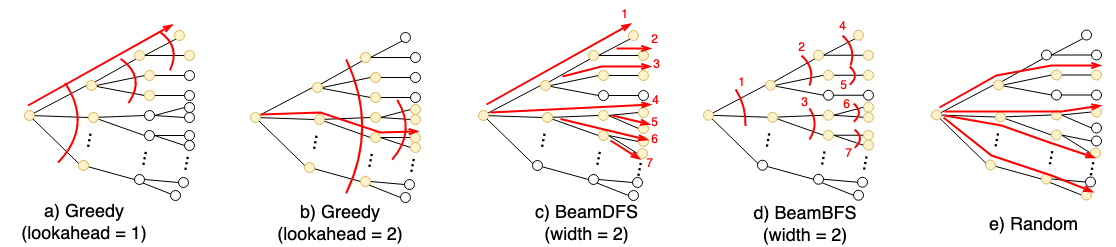}
\vspace*{-0.2in}
\caption{Traditional search approach in finding the optimal sequence. Actions (edges) are sorted by the performance of the next state.}
\label{fig/searches}
\end{figure*}

\section{Search to Optimize Tensor Programs}\label{sec/trad_approach}

The traditional approach for auto-tuning tensor programs is based on hill climbing, genetic, and various search algorithms \cite{ashouri2018survey}. These algorithms can find performant schedules for a single program, but the search time and the quality of the solution depend heavily on the smoothness of the optimization space. If the optimization sequence to highly rewarded states includes some actions that produce negative rewards, hill climbing algorithms can converge to local minima. Genetic algorithms, on the other hand, use a lot of computational resources, because they converge slowly.

We implemented the following set of search algorithms (Figure \ref{fig/searches}) to identify the difficulty of the problem:

\begin{itemize}
    \item Greedy search with lookahead of 1 and 2
    \vspace*{0.05in}
    \item Beam Depth First Search (BeamDFS) with width 2, 4
    \vspace*{0.05in}
    \item Beam Breath First Search (BeamBFS) with width 2, 4
    \vspace*{0.05in}
    \item Random search
    \vspace*{0.1in}
\end{itemize}

First, we introduce the family of Greedy search algorithms with arbitrary lookahead. In each step of this algorithm, we evaluate all possible states after applying lookahead steps and select the step toward the most promising state. With a lookahead of 1, the agent stops if there is no better action than the current state, while the lookahead of 2 enables the agent to tolerate one bad step that leads to a more promising solution. Ideally, with a large enough lookahead, Greedy Search would be able to overcome the problem of local minima for actions with negative rewards. Unfortunately, such computation comes with the cost of \(O(steps * |action\_space|^{lookahead})\), which is prohibitively expensive for large lookaheads.

Second, we implemented a family of Beam search algorithms with arbitrary width. In each step, we calculate the best width actions and expand them further until we reach the specified depth of the search tree. Expansion of the states could be done in depth-first (BeamDFS) and breadth-first (BeamBFS) manner and search properties drastically differ when search time elapses before the full search graph is constructed. Complexity of both of these approach is \(O(width^{steps})\), where \(width < |action\_space|\).

BeamDFS can be seen as an extension to the Greedy algorithm with a lookahead of 1, with few additions. It doesn't terminate if the next state is worse than the current and it recursively visits all states of the search graph where each node has maximum \textit{width} children. This enables it to tolerate non-convex parts of spaces as long as the optimal action ranks better than other actions in the current step.

BeamBFS variant finds iteratively a performant action sequence as it builds a search graph for each number of steps. This approach would be beneficial if the performant sequence is shorter than the specified search depth.

Finally, Random search randomly chooses a sequence of actions with a specified length. The benefit of this search is that it can uniformly explore a large number of diverse states providing a general idea about the landscape. From our experiments, random search provides surprisingly good results that we elaborate on in the next section.

\section{Evaluation}\label{sec/evaluation}

We evaluated LoopTune on a series of benchmarks to answer to following questions:

\begin{itemize}
    \item How do different RL algorithms compare to each other?\vspace*{0.05in}
    \item How does LoopTune compare to traditional search algorithms?\vspace*{0.05in}
    \item How does LoopTune compare to optimized libraries and auto-tuners like TVM?\vspace*{0.1in}
\end{itemize}
  
The benchmark dataset consists of synthesized loop nests for matrix multiplication. The matrix multiplication dataset has 2197 untiled loop nests for matrices with dimensions in the range from 64 to 256 with the step of 16. We chose these ranges since they are used as tile sizes for customized libraries such as cuBLAS \cite{mm_guide}. We train on the 80\% split of the dataset (size 1757) while leaving 20\% for the test dataset (size 440).

Experiments are performed on an Intel Xeon CPU running on 2.20GHz, with 40 CPU cores and 2 Nvidia Quadro GP100 GPUs. The CPU has cache sizes L1(data/instruction) 1.3 MB, L2 10MB, L3 52MB.

\begin{figure}[ht]
\setcounter{figure}{6}
\includegraphics[width=0.48\textwidth]{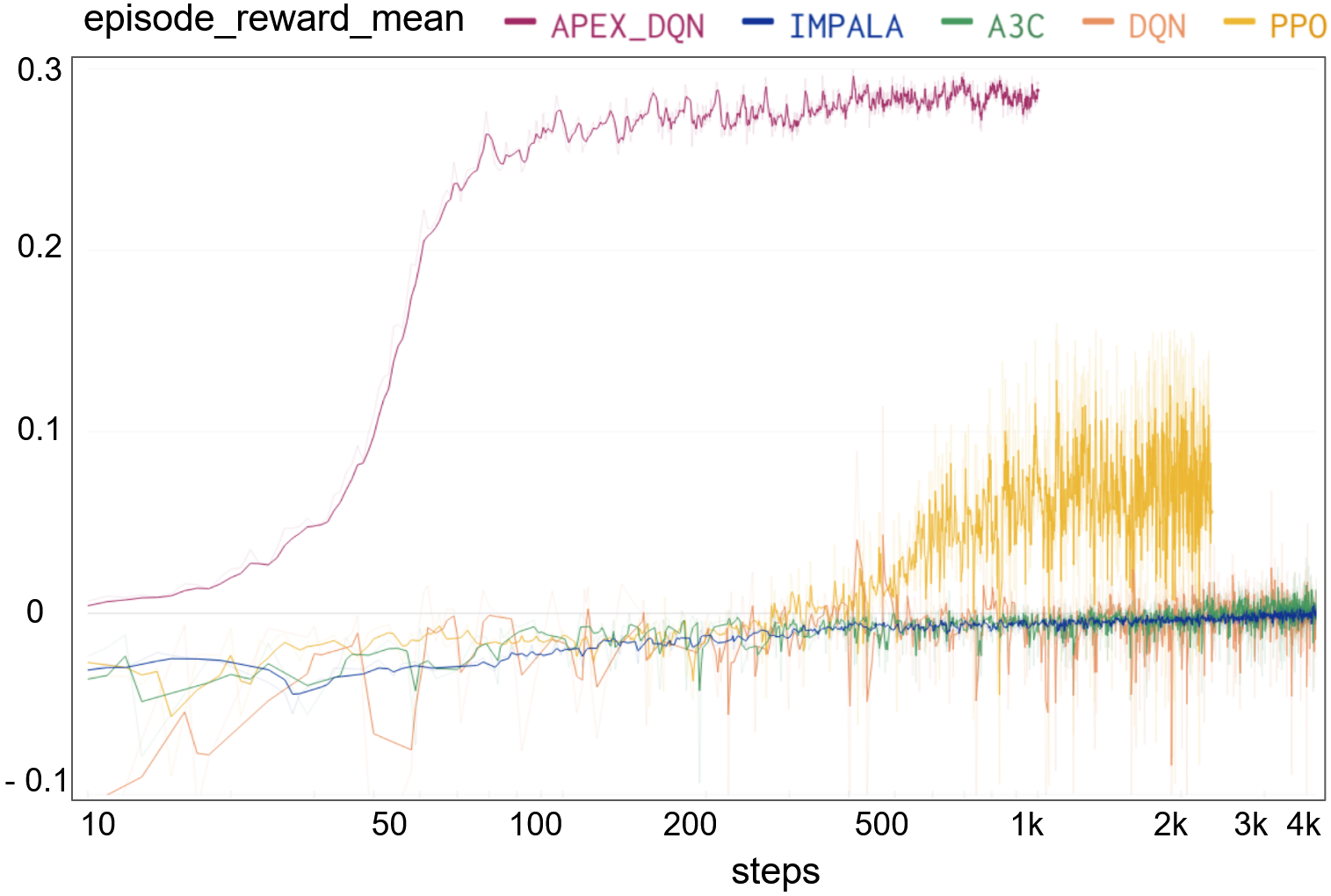}
% \vspace*{-0.2in}
\caption{Average reward per epoch for RLlib algorithms during training of 4000 steps.}
\label{fig/rllib_training}
\vspace{-0.2in}
\end{figure}

\subsection{RLlib Training Analysis}
To train LoopTune, we use Ray's RLlib library \cite{liang2018rllib}. After we define our environment in CompilerGym and register it with RLlib, we need to instantiate the appropriate trainer and find its most promising hyper-parameters. To find the best trainer we compare PPO, A3C, DQN, APEX\_DQN, and Impala. In all cases as a model, we use the network with fully connected layers, with arbitrary width and the number of layers, and the same feature representation. 

To find the optimal parameters for each trainer, we run a hyper-parameter sweep for the learning rate, exploration factor, depth, and width of the neural network. After finding the best parameters for each trainer, we run the final training for 4000 iterations and stop training early if the average reward per epoch converges. In each iteration, the optimizer applies the episode of 10 actions and updates the neural network. Finally, we compare trainers by plotting the \(episode\_reward\_mean\) which represents the averaged increase of GFLOPS achieved in the episode normalized to the peak performance of the device (Figure \ref{fig/rllib_training}).

We found that the APEX\_DQN trainer performs an order of magnitude better than other trainers, converging after roughly 200 steps and providing an average increase of 30\% of the peak performance (peak = 114.204 GFLOPS). In contrast, PPO required more than 1000 steps to converge to an improvement of 8\% of the peak, while Impala, A3C, and DQN have not been able to achieve positive results. We trained APEX\_DQN for 17.5 hours for 1000 iterations, which could be shortened to 3.5h with early stopping after 200 iterations.
% The hyper-parameters for the winning APEX\_DQN configuration include: lr = 1e6, gamma = 0.95, network\_depth = 10, network\_width = 1000.

We believe that the superiority of APEX\_DQN lies in the capability to prioritize the most significant experiences generated by the actors. Since all the algorithms are implemented in RLlib, it takes only one line of change to benefit from novel algorithms in the future. We further compare the APEX\_DQN solution with non-RL approaches.

\begin{figure*}[ht]
\includegraphics[width=\textwidth]{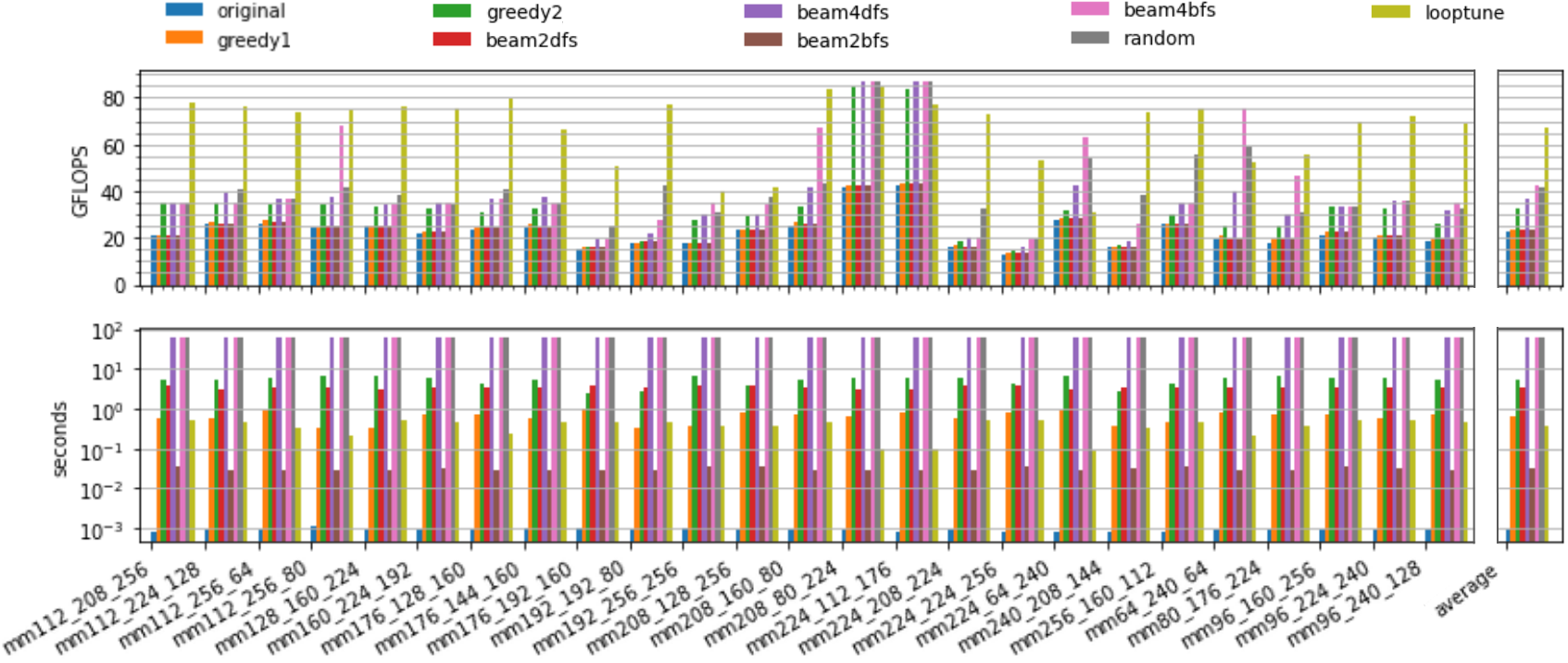}
\vspace*{-0.2in}
\caption{Achieved performance ({\bf higher} is better) and search time ({\bf lower} is better) of randomly selected 25 test benchmarks given 60 seconds for search. 
Original refers to LoopNest, which was used as a back compiler for greedy, beam, random searches, and the LoopTune method.}
\label{fig/test_bars}
\end{figure*}

\begin{figure}[ht!]
\includegraphics[width=0.45\textwidth]{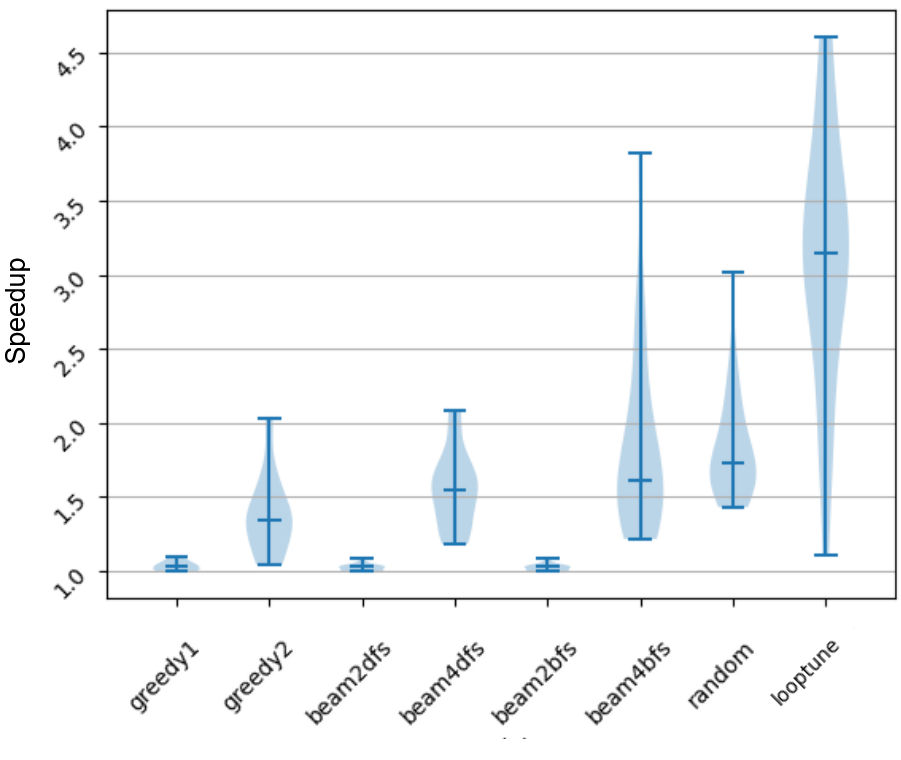}
\vspace*{-0.2in}
\caption{Speedup distribution for searches from Figure \ref{fig/test_bars} normalized with LoopNest results.}
\label{fig/violin}
% \vspace{-0.2in}
\end{figure}

\subsection{Comparison to Search Based Approaches}

To evaluate the difficulty of searching the optimization space, we run a set of traditional search algorithms including Greedy search with lookahead of 1 and 2, BFS and DFS variant of Beam search with widths 2 and 4, and Random search. We implemented each search with caching to avoid repeating evaluations of the same states. We run each search on a test dataset of 440 benchmarks, setting the time limit to 60 seconds. To compare traditional searches to policy generated from the RL approach, we visualize the search time and achieved performance of the produced code of 25 random benchmarks from the test set in Figure \ref{fig/test_bars}.

In 88\% test benchmarks, the APEX\_DQN policy network outperforms the best traditional searches by 1.8x on average in less than a second, which is an order of magnitude less time. To better understand the characteristics of each search we present the speedup distribution in Figure \ref{fig/violin}.

Increasing lookahead to 2 improves Greedy search's performance. Beam2BFS and Beam2DFS achieve poor results, despite exploring the entire search subtree, which implies that performant schedules include non-performant actions. Increasing the width to 4 significantly boosts performance, outperforming Greedy2. The success of Random search further emphasizes that optimization space is non-linear. Finally, RL policy network significantly outperforms all search methods by optimizing for long-range rewards up to 10 steps ahead, avoiding local minima.

\subsection{Analysis of the loop schedule optimization space}

\begin{figure}[h]
\includegraphics[width=0.48\textwidth]{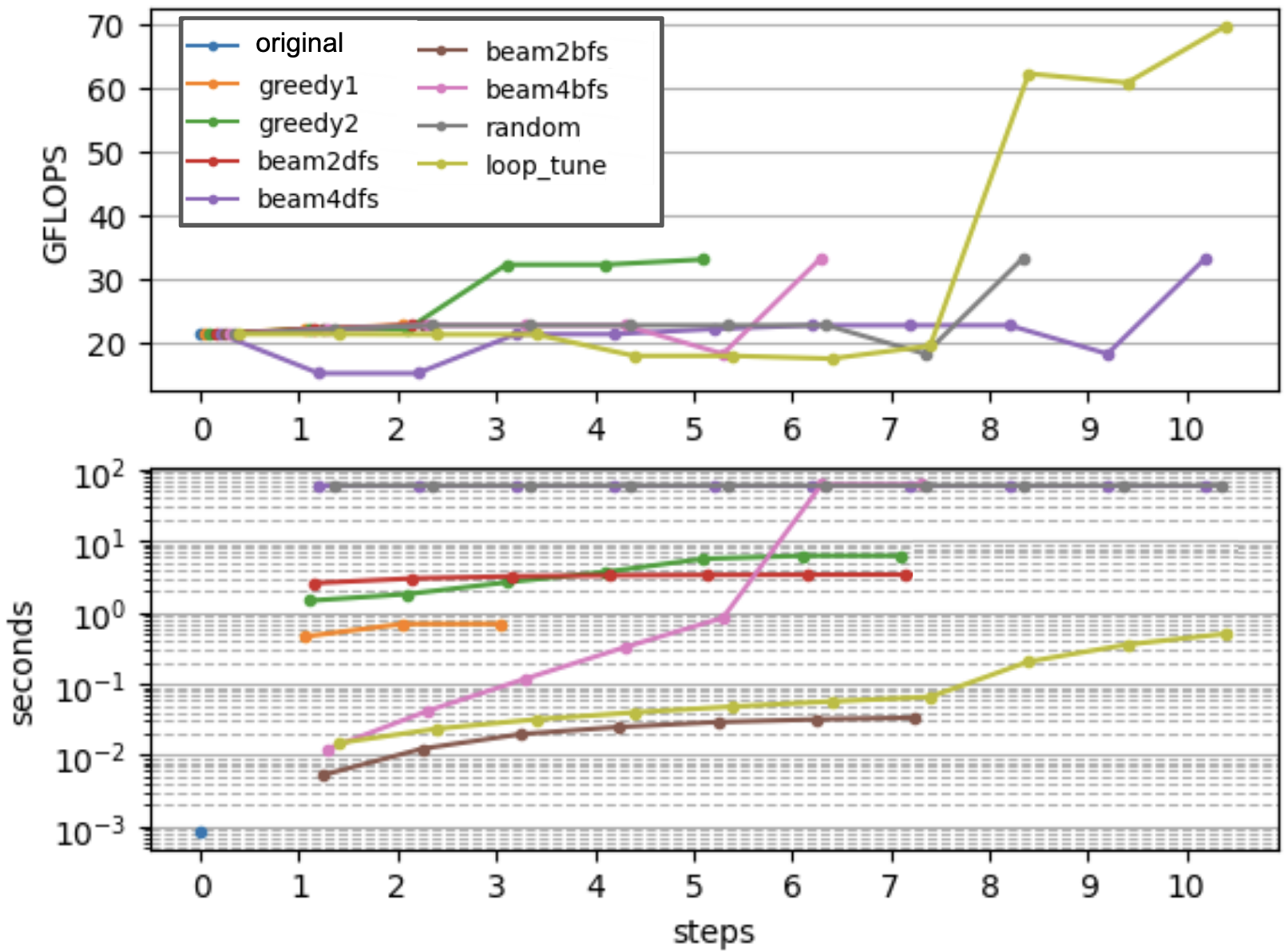}
\caption{Performance and time needed for expanding a search graph in each step.}
\label{fig/gain}
\end{figure}

Next, we visualize the performance and search speed of search algorithms and the RL approach for each step (Figure \ref{fig/gain}). The upper figure shows the reward signal in GFLOPS for the best-found schedule, while the lower figure shows how long it takes to choose an action for the given step. For the Depth-First search and Random search, actions are not decided until the end of the search graph construction and appear flat on the plot.

Greedy1 terminates quickly, creating a search graph of depth 2 and being stuck to the local minimum. Greedy2 can expand the graph up to depth 6, avoiding one-step local minima and achieving better performance, but still exploring only a small number of states. 

Beam2DFS expands the graph in-depth and each layer is updated during graph construction, keeping the time curve relatively flat. BeamBFS, on the other hand, builds the search space layer by layer completing lower layers first. 
The fact that Beam2DFS and Beam2BFS finished before the deadline (60s) means that they constructed the whole search graph of spawn 2. Neither of the two searches found a performant solution indicating that all performant schedules consist of at least one action which is best 2 actions.

Beam4DFS and Beam4BFS both terminated with a deadline which means that they only partially constructed their search graph of spawn 4. Beam4DFS's search graph includes solutions with long sequences of up to 10 steps, while Beam4BFS completely explored all solutions with 5 steps. In both cases, the best solutions contain long sequences of actions with non-monotonically increasing performance, which enables these searches to see further than Greedy searches. 

The Random search uses all time to expand the search graph from root to depth 10 without following any metric and evaluating each state in the graph. This way Random search can uniformly explore optimization space, including sequences of non-monotonic actions. 

RL policy network is able to outperform all previous algorithms by learning optimization patterns that maximize future rewards, speeding up the execution 3.2x times on average compared to the original LoopNest implementation. Its solution tolerates long sequences of non-performant actions, being worse than all other searches from the 4th to the 7th step to reach a performant state at the 8th step. Additionally, RL policy network search time grows linearly in the length of an action sequence, which enables us to use the policy network on harder problems that require a larger number of steps. These capabilities are paramount for auto-tuning general compilers such as LLVM.

\begin{figure*}[ht!]
\centering
\includegraphics[width=0.9\textwidth]{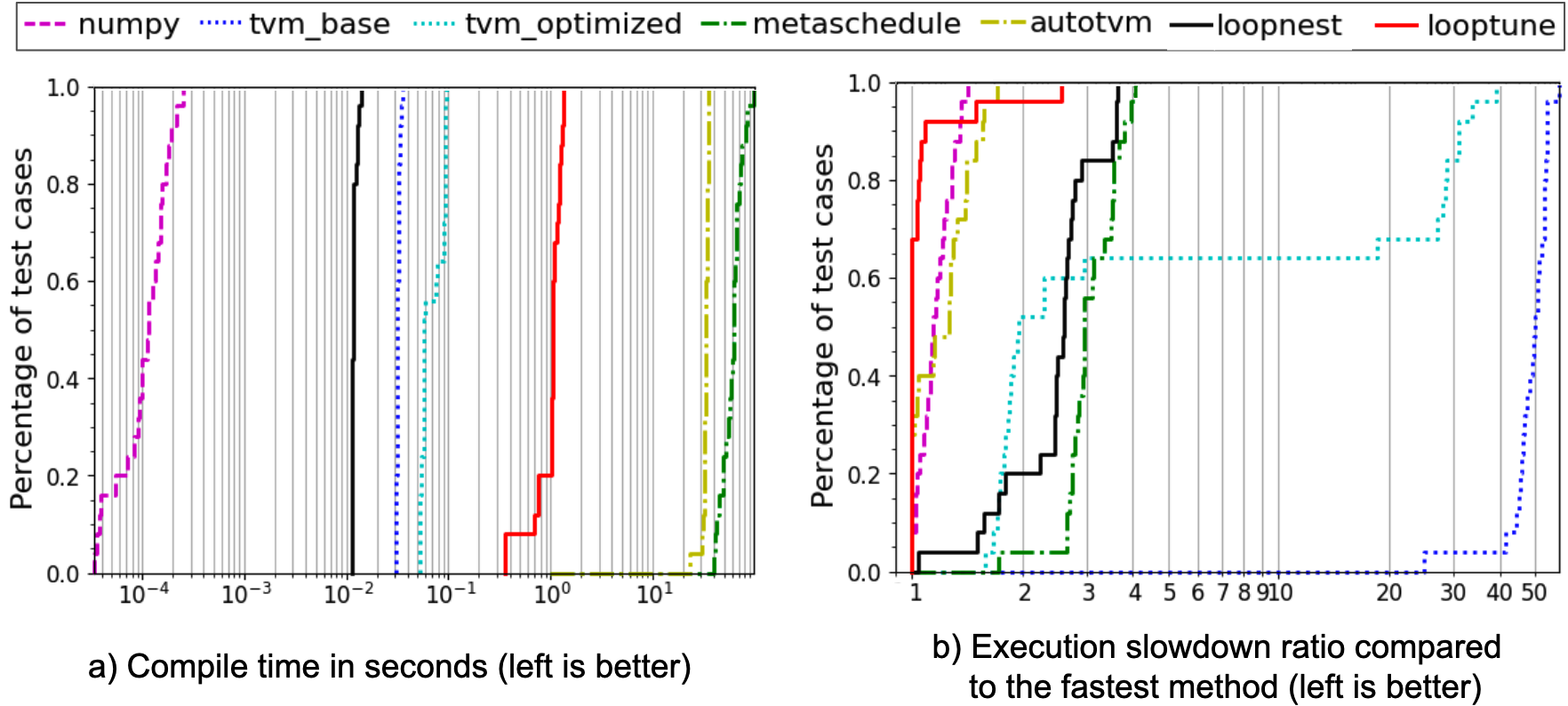}
% \vspace*{-0.2in}
\caption{ Compile time and Execution ratio of test benchmarks. For Figure b), test cases were normalized with the best method sorted from best to worst on the y-axis.}
\label{fig/tvm_bars}
\end{figure*}

\subsection{Comparison to Numpy, TVM, MetaSchedule, and AutoTVM}

Next, we show performance profiles \cite{dolan2002performanceprofile} for compilation and execution of generated code on the test dataset (440 examples), and compare it to a popular hand-tuned library for tensor operations -- Numpy\footnote{Numpy uses Intel’s state-of-the-art MKL implementation of BLAS.}, tensor compiler -- TVM (base version and optimized version with blocking, permutation, and vectorization) and widely used auto-tuners -- autoTVM and MetaSchedule (Figure \ref{fig/tvm_bars}). 

LoopTune outperforms all other approaches in 67\% of test cases while reaching more than 90\% of best performance in 92\% of test cases. On average LoopTune beats base TVM by 43x, optimized TVM by 9.7x, MetaSchedule by 2.8x, and AutoTVM by 1.08x, while being 3\% slower than Numpy.
In contrast to Numpy, LoopTune doesn't require hand-tuning which reduces development and maintenance costs. 

Moreover, LoopTune makes real-time autotuning practical, generating code in just 1 second, while autoTVM and MetaSchedule require 33 and 62 seconds on average. This is particularly important for applications that require downloading and tuning in real-time from web-based repositories. An example of this can be tuning image/video filters for social media apps and video games for mobile or VR devices. 

We used official documentation from TVM \cite{tvmdoc} to implement matrix multiplication for the examples from the test set. This implementation of TVM includes blocking, loop permutation, and vectorization optimizations, which are the same set of optimizations we are using for LoopTune. We enable the \textit{"llvm -mcpu=core-avx2"} option for TVM, MetaSchedule, and AutoTVM, to get the best results for our architecture. For MetaSchedule we used stochastic sampling, tiling, reordering, and unrolling, while for AutoTVM we used XGBTuner, evaluating 64 possible schedules for both. 

Evaluating more than 64 schedules would require proportionally more time, which makes it prohibitively long for our use case -- auto-tuning in a matter of seconds. For the same reason, we don't include in our evaluation popular cost-model based frameworks such as Ansor \cite{zheng2020ansor}, Value Learning \cite{steiner2021value} and TenSet \cite{zheng2021tenset} and FlexTensor \cite{zheng2020flextensor} since they have similar or longer search time.

\section{Related Work}\label{sec/related_work}

% General popular tensor compilers
\textbf{Tensor specific libraries.} Tensor-based mathematical notation was first used by APL \cite{abrams1970apl}. Similar to APL, modern frameworks such as NumPy \cite{van2011numpy}, Matlab's Tensor Toolbox \cite{bader2006algorithm}, Intel MKL \cite{mkl}, PyTorch \cite{pytorch} and Tensorflow \cite{abadi2016tensorflow} provide an intuitive interface for manipulating tensors, performing customized operations, and executing machine learning algorithms. Although these libraries often vectorize tensor computation, they hardly find the most performant order and sizes of the loop for custom hardware.

% % Domain-specific libraries
% Besides machine learning, tensor computations are used for quantum chemistry simulations. Libraries such as Tensor Contraction Engine \cite{di2014towards}, LibTensor \cite{epifanovsky2013new}, and Cyclops Tensor \cite{solomonik2014massively} provide general and sometimes domain-specific tensor computations. They often use distributed algorithms and tensor blocking. 

% Search-based autotuners
\textbf{Search-based compilers.} Rather than relying on one-size-fits-all solutions from expert-written libraries, projects ATLAS \cite{whaley1998automatically} and FFTW \cite{frigo1998fftw} empirically optimize BLAS and FFT routines given custom hardware. PetaBricks \cite{ansel2009petabricks} chooses the most appropriate algorithm of the computation for the given platform and tunes its parameter by using iterative methods. OpenTuner \cite{ansel2014opentuner} provides an ensemble of method-agnostic search techniques for program autotuning. Although these methods can be highly effective, auto-tuning requires significant search times for each program, which can be prohibitively expensive. LoopTune takes only a second to tune tensor computations.

\textbf{Graph-based compilers.} nGraph \cite{cyphers2018intel} passes its graph internal representation to a transformer and generates optimized code for the selected backend.
XLA \cite{sabne2020xla} automatically replaces subgraphs from Tensorflow with optimized binaries. Glow \cite{rotem2018glow} applies domain-specific optimization at a high level, memory-related optimizations at instruction-based intermediate representation, and hardware-specific optimization at the lowest level. MLIR \cite{lattner2021mlir} provides extensible compiler infrastructure that aims to unify domain-specific optimizers, providing multiple representations and layers of optimization. Rather than using a complex representation, LoopTune encodes graph-based representation to simple vectors with relevant features to describe memory access patterns. This enables fast inference with a simple multi-layer perceptron.

\textbf{Scheduling-based compilers.} Halide \cite{ragan2013halide} is the first influential work to propose the separation of computation and schedule for optimizing image processing and tensor computations. It uses a declarative language to specify tensor computations and a separate language for scheduling its execution. Similar to LoopTune, Halide's scheduling language includes operations such as splitting, and reordering, with the addition of vectorizing, unrolling, and parallelizing loops. TVM \cite{tvm} extends Halide's compute/schedule concept with hardware intrinsics and defines new optimizations such as tensorization and latency hiding. AutoTVM \cite{autotvm} extends the TVM cost model and adds a template-guided search framework. FlexTensor \cite{zheng2020flextensor} search directly schedule primitives on the finner-graned level than templates. In contrast to these approaches, LoopTune defines the action space with a policy model in mind eliminating parametrized actions that are hard to learn \cite{kanervisto2020action}.

% polyhedral compilers
\textbf{Polyhederal based compilers.} To represent tensor computation polyhedral optimizers Polly \cite{grosser2011polly} and others \cite{verdoolaege2010isl} \cite{bagnara2008parma} use linear programming and affine transformations to optimize loops static control-flow. Tensor comprehensions \cite{vasilache2018tensor} uses Halide's intermediate representation to represent computation, polyhedral representation to represent loops and just-in-time compilation for GPU.

\textbf{Cost-model based compilers.}
To speed up the evaluation of a computation, popular frameworks such as Ansor \cite{zheng2020ansor}, Value Learning \cite{steiner2021value}, and TenSet \cite{zheng2021tenset} learn a cost model to
evaluate performance and use decision trees, evolutionary search, and monte-carlo tree search to identify the best one. Although the performance cost model reduces evaluation time, converging to the optimal state in highly non-convex action spaces is difficult. Additionally, the inference with the cost model and basic greedy search requires \textit{actions\_sequence\_len * number\_of\_possible\_actions} inferences, while the policy network requires only \textit{actions\_sequence\_len} inferences which is exactly the case for LoopTune.

% Reinforcement learning based
\textbf{Policy-model based compilers.}
Neurovectorizer \cite{haj2020neurovectorizer} uses deep RL to improve the vectorization of CPU loops by tuning vectorization width and interleaving count. Chameleon \cite{ahn2020chameleon} uses a policy network to guide an adaptive sampling algorithm with domain knowledge to search configuration space. MLGO \cite{trofin2021mlgo} uses Policy gradient and Evolution strategies to optimize binary size by inlining functions. PolyGym \cite{brauckmann2021reinforcement} explores loop schedules combining polyhedral representation with RL and provides infrastructure for the user to apply different RL algorithms utilizing their representation. In contrast to these approaches, LoopTune proposes a novel graph-based representation, action space, and methodology for optimizing loop nests.
\section{Limitations and Future work}

One of LoopTune's limitations is that the loop nest shape needs to be known in compile time. For most of the ML computation, this is defined by design. Another limitation is that training time is proportional to the computation workload since we measure performance explicitly rather than using a cost model. For small kernels, this is not the problem, while for larger kernels it might be necessary to use a cost model during training. Finally, at the moment LoopTune has only support for CPU, although we looking forward to implement GPU support in the future.
% \newpage
\section{Conclusions}\label{sec/conclusion}

LoopTune is a novel performant auto-tuner for tensor computations on the CPU, capable of auto-tuning code in less than a second. LoopTune utilizes deep RL to train a policy network that reorders and tiles loop nests and applies hardware-specific optimization using LoopNest to tailor the loop nest to the underlying hardware. To map this problem to reinforcement learning, LoopTune introduces a unique action space, graph-based state representation, and reward signal.

By using RLlib’s APEX DQN algorithm, LoopTune speeds up the original LoopNest implementation by 3.2x given 1 second, on a suite of test problems, while the best traditional search algorithm achieved 1.8x given 60 seconds. LoopTune achieves an order of magnitude better results than the optimized implementation of TVM, which includes blocking, loop permutation, and vectorization. Additionally, LoopTune outperforms MetaSchedule and AutoTVM by 2.8x and 1.08x on average, generating code again in 1 second, while MetaSchedule and AutoTVM require 33 seconds and 62 seconds, respectively. This makes real-time auto-tuning possible.

Finally, LoopTune consistently performs at the same level as the expert-optimized library Numpy, significantly reducing development efforts. This finding further supports the belief that deep reinforcement learning techniques will play an important role in the next generation of compilers.

% \bibliographystyle{plain}
% \input{ref.tex}
% \bibliography{references.bib}
\bibliography{references.bib}

\begin{thebibliography}{10}

\bibitem{mm_guide}
Matrix multiplication background user's guide.
\newblock
  \url{https://docs.nvidia.com/deeplearning/performance/dl-performance-matrix-multiplication/index.html}.
\newblock Accessed: 2023-04-26.

\bibitem{abadi2016tensorflow}
Mart{\'\i}n Abadi, Paul Barham, Jianmin Chen, Zhifeng Chen, Andy Davis, Jeffrey
  Dean, Matthieu Devin, Sanjay Ghemawat, Geoffrey Irving, Michael Isard, et~al.
\newblock Tensorflow: A system for large-scale machine learning.
\newblock In {\em 12th $\{$USENIX$\}$ symposium on operating systems design and
  implementation ($\{$OSDI$\}$ 16)}, pages 265--283, 2016.

\bibitem{abdi2010pca}
Herv{\'e} Abdi and Lynne~J Williams.
\newblock Principal component analysis.
\newblock {\em Wiley interdisciplinary reviews: computational statistics},
  2(4):433--459, 2010.

\bibitem{abrams1970apl}
Philip~Samuel Abrams.
\newblock An apl machine.
\newblock Technical report, Stanford Linear Accelerator Center, Calif., 1970.

\bibitem{ahn2020chameleon}
Byung~Hoon Ahn, Prannoy Pilligundla, Amir Yazdanbakhsh, and Hadi Esmaeilzadeh.
\newblock Chameleon: Adaptive code optimization for expedited deep neural
  network compilation.
\newblock {\em arXiv preprint arXiv:2001.08743}, 2020.

\bibitem{albooyeh2019broadcast}
Marjan Albooyeh, Daniele Bertolini, and Siamak Ravanbakhsh.
\newblock Incidence networks for geometric deep learning.
\newblock {\em arXiv preprint arXiv:1905.11460}, 2019.

\bibitem{ansel2009petabricks}
Jason Ansel, Cy~Chan, Yee~Lok Wong, Marek Olszewski, Qin Zhao, Alan Edelman,
  and Saman Amarasinghe.
\newblock Petabricks: A language and compiler for algorithmic choice.
\newblock {\em ACM Sigplan Notices}, 44(6):38--49, 2009.

\bibitem{ansel2014opentuner}
Jason Ansel, Shoaib Kamil, Kalyan Veeramachaneni, Jonathan Ragan-Kelley,
  Jeffrey Bosboom, Una-May O'Reilly, and Saman Amarasinghe.
\newblock Opentuner: An extensible framework for program autotuning.
\newblock In {\em Proceedings of the 23rd international conference on Parallel
  architectures and compilation}, pages 303--316, 2014.

\bibitem{arm2016arm}
R~ARM.
\newblock Cortex-a57 software optimization guide.
\newblock {\em ARM}, 2016.

\bibitem{ashouri2018survey}
Amir~H Ashouri, William Killian, John Cavazos, Gianluca Palermo, and Cristina
  Silvano.
\newblock A survey on compiler autotuning using machine learning.
\newblock {\em ACM Computing Surveys (CSUR)}, 51(5):1--42, 2018.

\bibitem{bader2006algorithm}
Brett~W Bader and Tamara~G Kolda.
\newblock Algorithm 862: Matlab tensor classes for fast algorithm prototyping.
\newblock {\em ACM Transactions on Mathematical Software (TOMS)},
  32(4):635--653, 2006.

\bibitem{bagnara2008parma}
Roberto Bagnara, Patricia~M Hill, and Enea Zaffanella.
\newblock The parma polyhedra library: Toward a complete set of numerical
  abstractions for the analysis and verification of hardware and software
  systems.
\newblock {\em Science of Computer Programming}, 72(1-2):3--21, 2008.

\bibitem{brauckmann2021reinforcement}
Alexander Brauckmann, Andr{\'e}s Goens, and Jeronimo Castrillon.
\newblock A reinforcement learning environment for polyhedral optimizations.
\newblock {\em arXiv preprint arXiv:2104.13732}, 2021.

\bibitem{tvm}
Tianqi Chen, Thierry Moreau, Ziheng Jiang, Lianmin Zheng, Eddie Yan, Haichen
  Shen, Meghan Cowan, Leyuan Wang, Yuwei Hu, Luis Ceze, Carlos Guestrin, and
  Arvind Krishnamurthy.
\newblock {TVM}: An automated end-to-end optimizing compiler for deep learning.
\newblock In {\em 13th {USENIX} Symposium on Operating Systems Design and
  Implementation ({OSDI} 18)}, pages 578--594, Carlsbad, CA, October 2018.
  {USENIX} Association.

\bibitem{autotvm}
Tianqi Chen, Lianmin Zheng, Eddie Yan, Ziheng Jiang, Thierry Moreau, Luis Ceze,
  Carlos Guestrin, and Arvind Krishnamurthy.
\newblock Learning to optimize tensor programs.
\newblock In {\em Advances in Neural Information Processing Systems}, pages
  3389--3400, 2018.

\bibitem{chetlur2014cudnn}
Sharan Chetlur, Cliff Woolley, Philippe Vandermersch, Jonathan Cohen, John
  Tran, Bryan Catanzaro, and Evan Shelhamer.
\newblock cudnn: Efficient primitives for deep learning.
\newblock {\em arXiv preprint arXiv:1410.0759}, 2014.

\bibitem{choquette2021nvidia}
Jack Choquette, Wishwesh Gandhi, Olivier Giroux, Nick Stam, and Ronny
  Krashinsky.
\newblock Nvidia a100 tensor core gpu: Performance and innovation.
\newblock {\em IEEE Micro}, 41(2):29--35, 2021.

\bibitem{mkl}
Intel Corporation.
\newblock Mkl developer reference.
\newblock
  \url{https://software.intel.com/content/www/us/en/develop/documentation/mkl-developer-reference-c/top.html},
  2020.

\bibitem{onednn}
Intel Corporation.
\newblock Onednn.
\newblock \url{https://github.com/oneapi-src/oneDNN}, 2020.

\bibitem{cummins2022compilergym}
Chris Cummins, Bram Wasti, Jiadong Guo, Brandon Cui, Jason Ansel, Sahir Gomez,
  Somya Jain, Jia Liu, Olivier Teytaud, Benoit Steiner, et~al.
\newblock Compilergym: robust, performant compiler optimization environments
  for ai research.
\newblock In {\em 2022 IEEE/ACM International Symposium on Code Generation and
  Optimization (CGO)}, pages 92--105. IEEE, 2022.

\bibitem{cyphers2018intel}
Scott Cyphers, Arjun~K Bansal, Anahita Bhiwandiwalla, Jayaram Bobba, Matthew
  Brookhart, Avijit Chakraborty, Will Constable, Christian Convey, Leona Cook,
  Omar Kanawi, et~al.
\newblock Intel ngraph: An intermediate representation, compiler, and executor
  for deep learning.
\newblock {\em arXiv preprint arXiv:1801.08058}, 2018.

\bibitem{di2014towards}
Edoardo Di~Napoli, Diego Fabregat-Traver, Gregorio Quintana-Ort{\'\i}, and
  Paolo Bientinesi.
\newblock Towards an efficient use of the blas library for multilinear tensor
  contractions.
\newblock {\em Applied Mathematics and Computation}, 235:454--468, 2014.

\bibitem{tvmdoc}
TVM documentation version(0.11.dev0).
\newblock How to optimize gemm on cpu¶.
\newblock [Online; accessed 28-November-2022].

\bibitem{dolan2002performanceprofile}
Elizabeth~D Dolan and Jorge~J Mor{\'e}.
\newblock Benchmarking optimization software with performance profiles.
\newblock {\em Mathematical programming}, 91:201--213, 2002.

\bibitem{domagala2014tiling}
Lukasz Domagala, Fabrice Rastello, Sadayappan Ponnuswany, and Duco Van~Amstel.
\newblock A tiling perspective for register optimization.
\newblock {\em arXiv preprint arXiv:1406.0582}, 2014.

\bibitem{espeholt2018impala}
Lasse Espeholt, Hubert Soyer, Remi Munos, Karen Simonyan, Vlad Mnih, Tom Ward,
  Yotam Doron, Vlad Firoiu, Tim Harley, Iain Dunning, et~al.
\newblock Impala: Scalable distributed deep-rl with importance weighted
  actor-learner architectures.
\newblock In {\em International conference on machine learning}, pages
  1407--1416. PMLR, 2018.

\bibitem{frigo1998fftw}
Matteo Frigo and Steven~G Johnson.
\newblock Fftw: An adaptive software architecture for the fft.
\newblock In {\em Proceedings of the 1998 IEEE International Conference on
  Acoustics, Speech and Signal Processing, ICASSP'98 (Cat. No. 98CH36181)},
  volume~3, pages 1381--1384. IEEE, 1998.

\bibitem{xnnpack}
Google.
\newblock Xnnpack.
\newblock \url{https://github.com/google/XNNPACK}, 2020.

\bibitem{grosser2012polly}
Tobias Grosser, Armin Groesslinger, and Christian Lengauer.
\newblock Polly—performing polyhedral optimizations on a low-level
  intermediate representation.
\newblock {\em Parallel Processing Letters}, 22(04):1250010, 2012.

\bibitem{grosser2011polly}
Tobias Grosser, Hongbin Zheng, Raghesh Aloor, Andreas Simb{\"u}rger, Armin
  Gr{\"o}{\ss}linger, and Louis-No{\"e}l Pouchet.
\newblock Polly-polyhedral optimization in llvm.
\newblock In {\em Proceedings of the First International Workshop on Polyhedral
  Compilation Techniques (IMPACT)}, volume 2011, page~1, 2011.

\bibitem{haj2020neurovectorizer}
Ameer Haj-Ali, Nesreen~K Ahmed, Ted Willke, Yakun~Sophia Shao, Krste Asanovic,
  and Ion Stoica.
\newblock Neurovectorizer: End-to-end vectorization with deep reinforcement
  learning.
\newblock In {\em Proceedings of the 18th ACM/IEEE International Symposium on
  Code Generation and Optimization}, pages 242--255, 2020.

\bibitem{haj2019autophase}
Ameer Haj-Ali, Qijing Huang, William Moses, John Xiang, Ion Stoica, Krste
  Asanovic, and John Wawrzynek.
\newblock Autophase: Compiler phase-ordering for high level synthesis with deep
  reinforcement learning.
\newblock {\em arXiv preprint arXiv:1901.04615}, 2019.

\bibitem{horgan2018dqnapex}
Dan Horgan, John Quan, David Budden, Gabriel Barth-Maron, Matteo Hessel, Hado
  Van~Hasselt, and David Silver.
\newblock Distributed prioritized experience replay.
\newblock {\em arXiv preprint arXiv:1803.00933}, 2018.

\bibitem{intel2014intel}
R~Intel.
\newblock Intel 64 and ia-32 architectures optimization reference manual.
\newblock {\em Intel Corporation, Sept}, 2014.

\bibitem{jeong2012performance}
Hwancheol Jeong, Sunghoon Kim, Weonjong Lee, and Seok-Ho Myung.
\newblock Performance of sse and avx instruction sets.
\newblock {\em arXiv preprint arXiv:1211.0820}, 2012.

\bibitem{jia2019graphcore}
Zhe Jia, Blake Tillman, Marco Maggioni, and Daniele~Paolo Scarpazza.
\newblock Dissecting the graphcore ipu architecture via microbenchmarking.
\newblock {\em arXiv preprint arXiv:1912.03413}, 2019.

\bibitem{jouppi2017datacenter}
Norman~P Jouppi, Cliff Young, Nishant Patil, David Patterson, Gaurav Agrawal,
  Raminder Bajwa, Sarah Bates, Suresh Bhatia, Nan Boden, Al~Borchers, et~al.
\newblock In-datacenter performance analysis of a tensor processing unit.
\newblock In {\em Proceedings of the 44th annual international symposium on
  computer architecture}, pages 1--12, 2017.

\bibitem{kanervisto2020action}
Anssi Kanervisto, Christian Scheller, and Ville Hautam{\"a}ki.
\newblock Action space shaping in deep reinforcement learning.
\newblock In {\em 2020 IEEE Conference on Games (CoG)}, pages 479--486. IEEE,
  2020.

\bibitem{krizhevsky2017imagenet}
Alex Krizhevsky, Ilya Sutskever, and Geoffrey~E Hinton.
\newblock Imagenet classification with deep convolutional neural networks.
\newblock {\em Communications of the ACM}, 60(6):84--90, 2017.

\bibitem{lattner2004llvm}
Chris Lattner and Vikram Adve.
\newblock Llvm: A compilation framework for lifelong program analysis \&
  transformation.
\newblock In {\em International Symposium on Code Generation and Optimization,
  2004. CGO 2004.}, pages 75--86. IEEE, 2004.

\bibitem{lattner2021mlir}
Chris Lattner, Mehdi Amini, Uday Bondhugula, Albert Cohen, Andy Davis, Jacques
  Pienaar, River Riddle, Tatiana Shpeisman, Nicolas Vasilache, and Oleksandr
  Zinenko.
\newblock Mlir: Scaling compiler infrastructure for domain specific
  computation.
\newblock In {\em 2021 IEEE/ACM International Symposium on Code Generation and
  Optimization (CGO)}, pages 2--14. IEEE, 2021.

\bibitem{liang2018rllib}
Eric Liang, Richard Liaw, Robert Nishihara, Philipp Moritz, Roy Fox, Ken
  Goldberg, Joseph Gonzalez, Michael Jordan, and Ion Stoica.
\newblock Rllib: Abstractions for distributed reinforcement learning.
\newblock In {\em International Conference on Machine Learning}, pages
  3053--3062. PMLR, 2018.

\bibitem{lomont2011introduction}
Chris Lomont.
\newblock Introduction to intel advanced vector extensions.
\newblock {\em Intel white paper}, 23, 2011.

\bibitem{markidis2018nvidia}
Stefano Markidis, Steven~Wei Der~Chien, Erwin Laure, Ivy~Bo Peng, and Jeffrey~S
  Vetter.
\newblock Nvidia tensor core programmability, performance \& precision.
\newblock In Pat Langley, editor, {\em 2018 IEEE international parallel and
  distributed processing symposium workshops (IPDPSW)}, pages 522--531,
  Stanford, CA, 2018. IEEE.

\bibitem{matthews2018high}
Devin~A Matthews.
\newblock High-performance tensor contraction without transposition.
\newblock {\em SIAM Journal on Scientific Computing}, 40(1):C1--C24, 2018.

\bibitem{mnih2016a3c}
Volodymyr Mnih, Adria~Puigdomenech Badia, Mehdi Mirza, Alex Graves, Timothy
  Lillicrap, Tim Harley, David Silver, and Koray Kavukcuoglu.
\newblock Asynchronous methods for deep reinforcement learning.
\newblock In {\em International conference on machine learning}, pages
  1928--1937. PMLR, 2016.

\bibitem{mnih2013playing}
Volodymyr Mnih, Koray Kavukcuoglu, David Silver, Alex Graves, Ioannis
  Antonoglou, Daan Wierstra, and Martin Riedmiller.
\newblock Playing atari with deep reinforcement learning.
\newblock {\em arXiv preprint arXiv:1312.5602}, 2013.

\bibitem{mnih2013dqn}
Volodymyr Mnih, Koray Kavukcuoglu, David Silver, Alex Graves, Ioannis
  Antonoglou, Daan Wierstra, and Martin Riedmiller.
\newblock Playing atari with deep reinforcement learning.
\newblock {\em arXiv preprint arXiv:1312.5602}, 2013.

\bibitem{pytorch}
Adam Paszke, Sam Gross, Francisco Massa, Adam Lerer, James Bradbury, Gregory
  Chanan, Trevor Killeen, Zeming Lin, Natalia Gimelshein, Luca Antiga, Alban
  Desmaison, Andreas Kopf, Edward Yang, Zachary DeVito, Martin Raison, Alykhan
  Tejani, Sasank Chilamkurthy, Benoit Steiner, Lu~Fang, Junjie Bai, and Soumith
  Chintala.
\newblock Pytorch: An imperative style, high-performance deep learning library.
\newblock In H.~Wallach, H.~Larochelle, A.~Beygelzimer, F.~d\textquotesingle
  Alch\'{e}-Buc, E.~Fox, and R.~Garnett, editors, {\em Advances in Neural
  Information Processing Systems 32}, pages 8024--8035. Curran Associates,
  Inc., 2019.

\bibitem{radu2018multimodal}
Valentin Radu, Catherine Tong, Sourav Bhattacharya, Nicholas~D Lane, Cecilia
  Mascolo, Mahesh~K Marina, and Fahim Kawsar.
\newblock Multimodal deep learning for activity and context recognition.
\newblock {\em Proceedings of the ACM on Interactive, Mobile, Wearable and
  Ubiquitous Technologies}, 1(4):1--27, 2018.

\bibitem{ragan2013halide}
Jonathan Ragan-Kelley, Connelly Barnes, Andrew Adams, Sylvain Paris, Fr{\'e}do
  Durand, and Saman Amarasinghe.
\newblock Halide: a language and compiler for optimizing parallelism, locality,
  and recomputation in image processing pipelines.
\newblock {\em Acm Sigplan Notices}, 48(6):519--530, 2013.

\bibitem{rocki2020cerebras}
Kamil Rocki, Dirk Van~Essendelft, Ilya Sharapov, Robert Schreiber, Michael
  Morrison, Vladimir Kibardin, Andrey Portnoy, Jean~Francois Dietiker, Madhava
  Syamlal, and Michael James.
\newblock Fast stencil-code computation on a wafer-scale processor.
\newblock In {\em SC20: International Conference for High Performance
  Computing, Networking, Storage and Analysis}, pages 1--14. IEEE, 2020.

\bibitem{rotem2018glow}
Nadav Rotem, Jordan Fix, Saleem Abdulrasool, Garret Catron, Summer Deng, Roman
  Dzhabarov, Nick Gibson, James Hegeman, Meghan Lele, Roman Levenstein, et~al.
\newblock Glow: Graph lowering compiler techniques for neural networks.
\newblock {\em arXiv preprint arXiv:1805.00907}, 2018.

\bibitem{sabne2020xla}
Amit Sabne.
\newblock Xla : Compiling machine learning for peak performance, 2020.

\bibitem{schulman2017ppo}
John Schulman, Filip Wolski, Prafulla Dhariwal, Alec Radford, and Oleg Klimov.
\newblock Proximal policy optimization algorithms.
\newblock {\em arXiv preprint arXiv:1707.06347}, 2017.

\bibitem{silver2016mastering}
David Silver, Aja Huang, Chris~J Maddison, Arthur Guez, Laurent Sifre, George
  Van Den~Driessche, Julian Schrittwieser, Ioannis Antonoglou, Veda
  Panneershelvam, Marc Lanctot, et~al.
\newblock Mastering the game of go with deep neural networks and tree search.
\newblock {\em nature}, 529(7587):484--489, 2016.

\bibitem{stallman1999using}
Richard~M Stallman et~al.
\newblock {\em Using and porting the GNU compiler collection}, volume~86.
\newblock Free Software Foundation, 1999.

\bibitem{steiner2021value}
Benoit Steiner, Chris Cummins, Horace He, and Hugh Leather.
\newblock Value learning for throughput optimization of deep learning
  workloads.
\newblock {\em Proceedings of Machine Learning and Systems}, 3, 2021.

\bibitem{tekin2021state}
A~Tekin, A~Tuncer~Durak, C~Piechurski, D~Kaliszan, F~Aylin~Sungur,
  F~Roberts{\'e}n, and P~Gschwandtner.
\newblock State-of-the-art and trends for computing and interconnect network
  solutions for hpc and ai.
\newblock {\em Partnership for Advanced Computing in Europe, Available online
  at www. praceri. eu}, 2021.

\bibitem{trofin2021mlgo}
Mircea Trofin, Yundi Qian, Eugene Brevdo, Zinan Lin, Krzysztof Choromanski, and
  David Li.
\newblock Mlgo: a machine learning guided compiler optimizations framework.
\newblock {\em arXiv preprint arXiv:2101.04808}, 2021.

\bibitem{van2011numpy}
Stefan Van Der~Walt, S~Chris Colbert, and Gael Varoquaux.
\newblock The numpy array: a structure for efficient numerical computation.
\newblock {\em Computing in science \& engineering}, 13(2):22--30, 2011.

\bibitem{vasilache2018tensor}
Nicolas Vasilache, Oleksandr Zinenko, Theodoros Theodoridis, Priya Goyal,
  Zachary DeVito, William~S Moses, Sven Verdoolaege, Andrew Adams, and Albert
  Cohen.
\newblock Tensor comprehensions: Framework-agnostic high-performance machine
  learning abstractions.
\newblock {\em arXiv preprint arXiv:1802.04730}, 2018.

\bibitem{verdoolaege2010isl}
Sven Verdoolaege.
\newblock isl: An integer set library for the polyhedral model.
\newblock In {\em International Congress on Mathematical Software}, pages
  299--302. Springer, 2010.

\bibitem{wang2022automating}
Huanting Wang, Zhanyong Tang, Cheng Zhang, Jiaqi Zhao, Chris Cummins, Hugh
  Leather, and Zheng Wang.
\newblock Automating reinforcement learning architecture design for code
  optimization.
\newblock In {\em Proceedings of the 31st ACM SIGPLAN International Conference
  on Compiler Construction}, pages 129--143, 2022.

\bibitem{wasti2022loopstack}
Bram Wasti, Jos{\'e}~Pablo Cambronero, Benoit Steiner, Hugh Leather, and
  Aleksandar Zlateski.
\newblock Loopstack: a lightweight tensor algebra compiler stack.
\newblock {\em arXiv preprint arXiv:2205.00618}, 2022.

\bibitem{whaley1998automatically}
R~Clinton Whaley and Jack~J Dongarra.
\newblock Automatically tuned linear algebra software.
\newblock In {\em SC'98: Proceedings of the 1998 ACM/IEEE conference on
  Supercomputing}, pages 38--38. IEEE, 1998.

\bibitem{wittmann2015short}
Markus Wittmann, Thomas Zeiser, Georg Hager, and Gerhard Wellein.
\newblock Short note on costs of floating point operations on current x86-64
  architectures: Denormals, overflow, underflow, and division by zero.
\newblock {\em arXiv preprint arXiv:1506.03997}, 2015.

\bibitem{zheng2020ansor}
Lianmin Zheng, Chengfan Jia, Minmin Sun, Zhao Wu, Cody~Hao Yu, Ameer Haj-Ali,
  Yida Wang, Jun Yang, Danyang Zhuo, Koushik Sen, et~al.
\newblock Ansor: Generating high-performance tensor programs for deep learning.
\newblock {\em arXiv preprint arXiv:2006.06762}, 2020.

\bibitem{zheng2021tenset}
Lianmin Zheng, Ruochen Liu, Junru Shao, Tianqi Chen, Joseph~E Gonzalez, Ion
  Stoica, and Ameer~Haj Ali.
\newblock Tenset: A large-scale program performance dataset for learned tensor
  compilers.
\newblock In {\em Thirty-fifth Conference on Neural Information Processing
  Systems Datasets and Benchmarks Track (Round 1)}, 2021.

\bibitem{zheng2020flextensor}
Size Zheng, Yun Liang, Shuo Wang, Renze Chen, and Kaiwen Sheng.
\newblock Flextensor: An automatic schedule exploration and optimization
  framework for tensor computation on heterogeneous system.
\newblock In {\em Proceedings of the Twenty-Fifth International Conference on
  Architectural Support for Programming Languages and Operating Systems}, pages
  859--873, 2020.

\end{thebibliography}

% \bibliography{example_paper}
% \bibliographystyle{icml2023}
\bibliographystyle{plain}

%%%%%%%%%%%%%%%%%%%%%%%%%%%%%%%%%%%%%%%%%%%%%%%%%%%%%%%%%%%%%%%%%%%%%%%%%%%%%%%
%%%%%%%%%%%%%%%%%%%%%%%%%%%%%%%%%%%%%%%%%%%%%%%%%%%%%%%%%%%%%%%%%%%%%%%%%%%%%%%
% APPENDIX
%%%%%%%%%%%%%%%%%%%%%%%%%%%%%%%%%%%%%%%%%%%%%%%%%%%%%%%%%%%%%%%%%%%%%%%%%%%%%%%
%%%%%%%%%%%%%%%%%%%%%%%%%%%%%%%%%%%%%%%%%%%%%%%%%%%%%%%%%%%%%%%%%%%%%%%%%%%%%%%
\newpage

% \appendix
% \input{loopnest}
% \input{appendix_trad_opt}

%%%%%%%%%%%%%%%%%%%%%%%%%%%%%%%%%%%%%%%%%%%%%%%%%%%%%%%%%%%%%%%%%%%%%%%%%%%%%%%
%%%%%%%%%%%%%%%%%%%%%%%%%%%%%%%%%%%%%%%%%%%%%%%%%%%%%%%%%%%%%%%%%%%%%%%%%%%%%%%

\end{document}